\documentclass{article}

\usepackage[main, final]{neurips_2026}

\usepackage[utf8]{inputenc} 
\usepackage[T1]{fontenc}    
\usepackage{hyperref}       
\usepackage{url}            
\usepackage{booktabs}       
\usepackage{amsfonts}       
\usepackage{nicefrac}       
\usepackage{microtype}      
\usepackage{xcolor}         
\usepackage{hyperref}
\usepackage{xspace}
\usepackage{multirow}
\usepackage{tabulary}
\usepackage{amsmath}
\usepackage[most]{tcolorbox}
\usepackage{enumitem}
\usepackage{microtype}
\usepackage{graphicx}
\usepackage{subcaption}
\usepackage{booktabs} 
\usepackage{pifont}

\usepackage{hyperref}
\usepackage{xspace}
\usepackage{multirow}
\usepackage{tabulary}
\usepackage{amsmath}
\usepackage[most]{tcolorbox}
\usepackage{enumitem}
\newenvironment{myenumerate}
    {\begin{enumerate}[label=(\arabic*),
        left=0pt,
        itemsep=5pt,
        parsep=0em,
        topsep=0.1em,
        partopsep=0.1em,
        listparindent=\parindent]}
    {\end{enumerate}}

\usepackage{xcolor}
\definecolor{gradstart}{HTML}{7B8FF7}
\definecolor{gradend}{HTML}{D84CB5}

\newcommand{\sys}{\textsc{NeuroRVQ}\xspace}

\newcommand{\neurorvq}{%
  \textcolor{gradstart!100!gradend}{N}%
  \textcolor{gradstart!85!gradend}{e}%
  \textcolor{gradstart!71!gradend}{u}%
  \textcolor{gradstart!57!gradend}{r}%
  \textcolor{gradstart!43!gradend}{o}%
  \textcolor{gradstart!28!gradend}{R}%
  \textcolor{gradstart!14!gradend}{V}%
  \textcolor{gradstart!0!gradend}{Q}%
}

\title{%
  \raisebox{-0.15\height}{\includegraphics[height=1.2em]{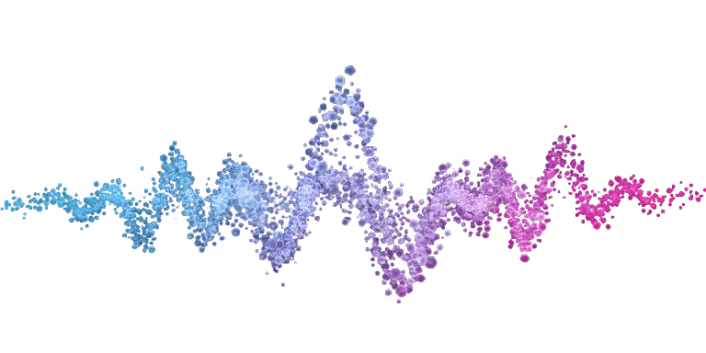}}%
  \hspace{0.3em}%
  {\sffamily\neurorvq}: Multi-Scale Biosignal Tokenization for Generative Foundation Models%
}

\author{
Konstantinos Barmpas \textsuperscript{*} \\
Imperial College London \& Cogitat \\
\And
Na Lee \\
Imperial College London \& Cogitat \\
\AND
Dimitrios Chalatsis \\
Imperial College London \\
\And
William Raftery \\
Imperial College London \\
\And
Yannis Panagakis \\
National and Kapodistrian University of Athens \& \\Archimedes Research Unit \& Cogitat \\
\AND
Dimitrios A. Adamos \\
Imperial College London \& Cogitat \\
\And
Nikolaos Laskaris \\
Aristotle University of Thessaloniki \& Cogitat \\
\And
Alexandros Koliousis\\
Northeastern University London \\
\And
Dario Farina \\
Imperial College London \\
\And
Stefanos Zafeiriou \\
Imperial College London \& Cogitat \\
}

\begin{document}

\maketitle

\renewcommand{\thefootnote}{\fnsymbol{footnote}} 
\footnotetext[1]{Correspondence to konstantinos.barmpas16@imperial.ac.uk.}

\begin{abstract}
Biosignals such as electroencephalography (EEG), electrocardiography (ECG), and electromyography (EMG) encode physiological activity across multiple temporal and spectral scales, yielding representations that are rich but challenging for machine learning. Foundation models trained to predict masked signal tokens have shown promise in learning generalizable biosignal representations, yet their performance depends on the tokenizer's ability to preserve high-frequency dynamics and reconstruct signals with high fidelity. We introduce \sys, a modality-adaptive biosignal tokenizer family designed for high-fidelity signal reconstruction. To capture the full frequency spectrum, \sys decomposes biosignals into frequency-specific representations via multi-scale temporal convolutions, each encoded into hierarchical RVQ codebooks to preserve high-frequency detail, combined with a novel phase-aware training loss that respects the circular topology of Fourier phase. By tuning the temporal resolution, number and size of temporal kernels and RVQ depth, this design adapts to the spectro-temporal characteristics of each biosignal modality. To validate that tokenizer quality drives downstream performance, we train a simple masked-token foundation model for each modality (\sys-FM) using the corresponding \sys tokenizer. The \sys-FM family achieves competitive or superior downstream performance compared to existing modality-specific foundation models, demonstrating that high-fidelity tokenization is a critical factor for effective biosignal modeling. Code and models are available on our project page: \href{https://neurorvq.github.io}{https://neurorvq.github.io}
\end{abstract}

\renewcommand{\thefootnote}{\arabic{footnote}}
\setcounter{footnote}{0}

\section{Introduction}

Biosignals, electrical recordings of physiological activity, are fundamental to modern healthcare and Human-Computer Interaction (HCI): Electroencephalography (EEG) captures brainwave dynamics underlying cognition, sleep and emotion; electrocardiography (ECG) records the cardiac electrical cycle critical for diagnosing arrhythmias and heart disease; and electromyography (EMG) measures muscular activation patterns essential for movement analysis and prosthetic control. Across these modalities, accurate signal decoding underpins a broad spectrum of applications, from Brain-Computer Interfaces (BCIs) for emotion recognition~\citep{bci_emotion_ref1, bci_emotion_ref2} and epileptic seizure detection~\citep{bci_epileptic_ref1, bci_epileptic_ref2}, to cardiac anomaly detection~\citep{hannun2019cardiologist} and gesture recognition from muscle activity~\citep{atzori2014electromyography}.
Despite their different physiological origins, EEG, ECG and EMG signals share a common characteristic: \textbf{information is distributed across multiple frequency scales.} EEG brainwaves span distinct frequency bands (e.g., delta, alpha, gamma) that capture various physiological states~\citep{Newson2018_lm}. ECG signals encode both low-frequency morphological features (e.g., the QRS complex) and subtle high-frequency components indicative of abnormalities ~\citep{goldberger2017clinical}. Similarly, EMG signals combine low-frequency motor unit recruitment patterns with high-frequency firing dynamics~\citep{deluca1997use}. \textbf{Any analysis framework for biosignals must capture this multiscale structure, both slow and fast dynamics, to fully preserve the information content of the signal.}

A range of techniques, from advanced signal processing to deep learning, have been employed for biosignal decoding. More recently, foundation models trained with self-supervised masked learning have emerged as a powerful paradigm: signals are first discretized into \emph{tokens} by a tokenizer and during training, tokens are intentionally masked so that the model learns to reconstruct them. This approach has shown strong generalization capabilities without relying on task-specific data collection. However, the success of masked modeling relies heavily on tokenizer quality. Tokenizers must effectively capture spatio-temporal patterns to convert continuous signals into meaningful discrete representations while enabling faithful reconstruction. \textbf{Current tokenizers struggle to preserve complete structural information, especially high-frequency components, necessary for robust generative masked modeling.}
The key to unlocking foundation-scale masked modeling for biosignals lies in the tokenizer:
\begin{tcolorbox}[colback=blue!5!white,colframe=cyan!75!black]
\centering
A well-designed biosignal tokenizer should not only compress continuous physiological signals into discrete neural tokens but also enable faithful reconstruction of the original waveform across all important frequency scales.
\end{tcolorbox}
In this work, we introduce \sys, a modality-adaptive biosignal tokenizer family designed for high-fidelity signal reconstruction. The main contributions of this work are:
\begin{myenumerate}
\item a \textbf{modality-adaptive multiscale tokenizer architecture} that decomposes biosignals into frequency-specific representations via temporal convolutions with varying kernel sizes and encodes each frequency scale into hierarchical Residual Vector Quantization (RVQ) codebooks, enabling high-fidelity signal reconstruction across EEG, ECG, and EMG. By tuning the temporal resolution, number and size of temporal kernels, and RVQ depth, this architecture adapts to the spectro-temporal characteristics of each biosignal modality.
\item a \textbf{phase-aware training loss} that reconstructs the Fourier spectrum through three complementary components: a log-amplitude loss for Fourier amplitude to emphasize high-frequency content, a temporal-domain regularization loss and a novel phase loss that respects the circular topology of phase angles by leveraging cosine similarity for directional alignment; and
\item a \textbf{simple masked-token foundation model} (\sys-FM) for each modality that leverages the corresponding \sys tokenizer during pre-training and achieves competitive or superior downstream performance compared to existing modality-specific foundation models, demonstrating that high-fidelity tokenization (rather than model scale or architectural complexity) is a critical ingredient for effective biosignal modeling.
\end{myenumerate}

Compared to existing approaches that operate at a single scale with a single codebook, \sys with its multi-scale architecture and per-scale hierarchical RVQ codebooks achieves substantially lower reconstruction error, while the phase-aware loss delivers the largest reconstruction improvement. When used in the pre-training of \sys-FM, the high-fidelity tokenization translates directly into strong downstream performance: \sys-EEG achieves the best mean balanced accuracy across five classification tasks, \sys-ECG doubles the next-best balanced accuracy on 43-class PTB-XL classification and \sys-EMG outperforms existing EMG foundation models across four classification benchmarks.

\section{Background \& Related Work}

The roots of biosignal analysis can be traced back to classical methods, when handcrafted features such as Power Spectral Density (PSD) and Independent Component Analysis (ICA) were regarded  standard~\citep{James_2005, Classical}. Although these methods provided valuable insights into biosignal decoding, they often failed to generalize due to high inter-subject variability and their limited capacity to capture the full complexity of brain, heart and muscle activities.


The deep learning era marked a shift towards data-driven feature extraction: models such as EEGNet~\citep{EEGNet}, EEGInception~\citep{EEGInception},  Brainwave-Scattering Net~\citep{Barmpas_Scattering} and EEGConformer~\citep{EEGConformer} for EEG, along with deep residual networks for ECG~\citep{Ribeiro2020-rv} and convolutional architectures for EMG~\citep{8630679}, reduced reliance on handcrafted features by automatically learning signal representations. To address domain-specific challenges such as subject variability, researchers incorporated techniques like subject-alignment into these architectures~\citep{ourNeurIPS2021, BEETL, Barmpas_2024_causality_jne}. Despite these advances, most deep  models remain limited: they require carefully labeled data and extensive task-specific training, hindering their ability to generalize across diverse tasks.



More recently, the success of large foundation models in natural language processing~\citep{brown2020language, touvron2023llama} and computer vision~\citep{mizrahi2023m} has motivated their adoption in biosignal decoding. While foundation models for EEG are now relatively mature, those for ECG and EMG are still emerging~\citep{Lee2025_survey}. Notable examples include BIOT~\citep{NEURIPS2023_f6b30f3e}, LaBraM~\citep{LaBraM}, NeuroGPT~\citep{NeuroGPT}, EEGPT~\citep{wang2024eegpt}, CBraMod~\citep{wang2025cbramod}, BrainOmni~\citep{xiao2025brainomni} and REVE~\citep{ouahidi2025reve} for EEG; HuBERT-ECG~\citep{HuBERT} and ECGFounder~\citep{ECGFounder} for ECG; and PhysioWave~\citep{chen2025physiowave} and TinyMyo~\citep{fasulo2026tinymyotinyfoundationmodel} for EMG. These models are pretrained on large unlabeled datasets using self-supervised signal reconstruction objectives and have shown promising generalization capabilities. However, they share a common bottleneck: the quality of signal tokenization, which determines the fidelity of the representations the model learns from.

Tokenization is critical for robust generative masked modeling, yet current tokenizers struggle to capture the full structure of biosignals. Some models rely on deep backbone-based tokenizers (e.g. CBraMod), while others employ transformer-based approaches (e.g. BIOT), which often learn entangled and redundant representations not optimized for reconstruction. As a result, the generated tokens lack the fidelity required for effective masked prediction. More recently, codebook-based approaches, such as LaBraM and BrainOmni, have been introduced to learn reconstructable discrete tokens using vector quantization (VQ) and residual VQ (RVQ) respectively. However, both rely on a single codebook to capture the full spectrum of biosignal activity and train their tokenizers with losses that do not always respect the signal processing properties of their reconstruction targets. The result is codebook-based tokenizers with unfaithful signal reconstruction.

\section{NeuroRVQ Tokenizer Design}

\begin{figure*}[t]
    \centering
    \includegraphics[width=\linewidth]{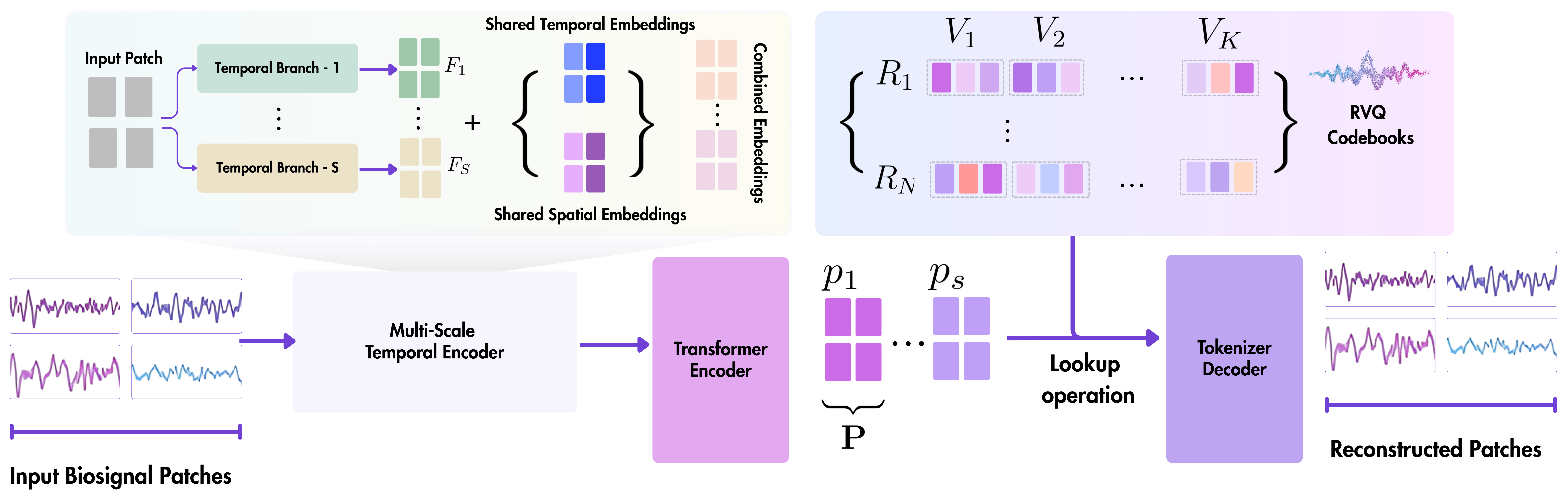}
     \caption{{\small The {\sffamily\neurorvq} tokenizer model. The multi-scale temporal encoder extracts multi-scale temporal features that are passed to the transformer encoder. RVQ codebooks per temporal branch discretize the biosignal patches into neural tokens. The tokenizer decoder reconstructs the biosignal by using the Fourier spectrum.}}
    \label{fig:neurorvq_tokenizer}
\end{figure*}

\subsection{Model Architecture}
\label{sec:model_arch_tokenizer}

\sys tokenizer is a specialized network designed to convert  raw biosignals into a sequence of compact and informative neural tokens. This transformation reduces the inherently high-dimensional and noisy nature of biosignals into a structured lower-dimensional representation preserving essential temporal–spectral patterns. In doing so, the tokenizer provides a kind of "grammar" for the underlying physiological activity.

Figure \ref{fig:neurorvq_tokenizer} illustrates the tokenizer architecture and processing flow. 
The input multi-variate biosignal timeseries is first segmented into patches. These patches are encoded by the multi-scale temporal encoder, that captures features in multiple resolutions and are then combined via the transformer encoder. For each scale, RVQ codebooks discretize the embeddings into a sequence of neural tokens. Finally, these tokens from all scales are combined and passed through the tokenizer decoder to reconstruct the input biosignal patches using the Fourier spectrum.

\paragraph{Generating patches.} Let $X \in \mathbb{R}^{C \times T}$ denote the input multi-variate biosignal, where $T$ is the number of time points and $C$ is the number of electrodes. The signal is first segmented into temporal patches. To ensure that the tokenizer can accommodate biosignals with arbitrary numbers of channels and variable durations, we adopt the following approach: during model pre-training, each input multi-variate biosignal is represented by $P$ patches of length $w$. This results in a segmented input sample
$\mathbf{x} \in \mathbb{R}^{P \times w}$ which provides flexibility to handle heterogeneous biosignal recordings and facilitates robust pre-training across diverse datasets.

\textbf{Multi-scale temporal encoder.} The first step is to extract multi-scale temporal features from each input patch using an inception-style module with $S$ distinct temporal scales. To achieve this, we apply 1-D temporal convolutions with varying kernel sizes $K_{\text{temporal}_1}, K_{\text{temporal}_2}, \dots, K_{\text{temporal}_{S}}$. Each temporal branch consists of the following sequence of operations: a 1-D convolution, followed by group normalization, GELU activation and pooling (this sequence of operations is repeated twice). This design enables the network to capture temporal features at multiple scales resulting in $S$ outputs $F_{1}, F_{2}, \dots, F_{S}$ where $F_{i} \in \mathbb{R}^{w}$ where $w$ is the patch time window length. 

\textbf{Transformer encoder.} Since the tokenization of the biosignal takes place in 
a per-patch level, the vital temporal and spatial information is incorporated through the use of trainable temporal $TE$ and spatial $SE$ embeddings, each with embedding dimension $w$. $SE = \{se_1,...,se_{|\mathcal{C}|}\}$ is indexed by the positions of the patches' electrodes within a list of all electrodes in the entire pre-training database $\mathcal{C}$ (e.g., 10-20 system for EEG or standard 12-lead for ECG) and $TE = \{te_1,...,te_P\}$ since each input consists of $P$ patches of length $w$.
These embeddings are added to the extracted multi-scale features $F_{1}, F_{2}, \dots, F_{S}$ and are shared across all $S$ temporal branches. The combined embeddings of each temporal branch are then passed through a series of shared transformer layers~\citep{vaswani2023attentionneed} resulting in the multi-scale patch representations $\mathbf{p}_{1}, \mathbf{p}_{2}, \dots, \mathbf{p}_{S} \in \mathbb{R}^{D}$. The shared transformer layers have incorporated the modification of \citet{dehghani23a}.

\textbf{RVQ codebook.} To efficiently discretize the multi-scale patch representations $\mathbf{p}_{1}, \mathbf{p}_{2}, \dots, \mathbf{p}_{S}$ while preserving fine-grained information of the biosignal, we use $S$ Residual Vector Quantization (RVQ) codebooks~\citep{Gray1984VectorQ}. For each temporal branch, the RVQ codebook $\mathcal{R}$ is defined as:

\vspace{-0.1in}
\begin{equation}
     \mathcal{R}  = \{\mathcal{V}_{i} | i= 1,...,N\}
\end{equation}

where $\mathcal{V}_{i}$ is a single codebook of neural tokens defined as:

\vspace{-0.1in}
\begin{equation}
     \mathcal{V}_{i}  = \{\mathbf{v}_{j} | j= 1,...,K\} \in \mathbb{R}^{K \times D}
\end{equation}

where $K$ is the number of discrete neural tokens and $D$ is the dimensionality of each token. The discretization of the patch representations is performed iteratively. For each $\mathcal{R}$, at step 1, a quantizer maps each patch representation $\mathbf{p}^{1}$ to its corresponding neural token $\mathbf{v}^{1} \in \mathcal{V}_{1}$. Specifically, for each patch, the quantizer selects the nearest neighbor $\mathbf{z_{1}}$ from the codebook $\mathcal{V}_{1}$:

\vspace{-0.1in}
\begin{equation}
    \mathbf{z_{1}} = \arg\min_{\mathbf{v^{1}} \in \mathcal{V}_{1}} \left\| l_{2}(\mathbf{p}^{1}) - l_{2}(\mathbf{v^{1}}) \right\|
\end{equation}

At each next step $i+1$, the residual patch representation $\mathbf{p}^{i+1}$ for the next codebook $\mathcal{V}_{i+1}$ is updated by subtracting the selected token $\mathbf{z}_{i}$:

\vspace{-0.2in}
\begin{equation}
    \mathbf{p}^{i+1} = \mathbf{p}^{i} - \mathbf{z}_{i}
\end{equation}

This process is repeated for all $N$ codebooks, producing a sequence of tokens $\mathbf{z}_1, \mathbf{z}_2, \dots, \mathbf{z}_N$ for each temporal branch separately. The final reconstructed patch representations $\hat{\mathbf{p}}$ is obtained by summing the contributions of all selected tokens $\mathbf{z}_1, \mathbf{z}_2, \dots, \mathbf{z}_N$:

\vspace{-0.1in}
\begin{equation}
    \hat{\mathbf{p}} = \sum_{i=1}^{N} \mathbf{z}_i
\end{equation}

At the end of this module, we have $\hat{\mathbf{p}}_1, \hat{\mathbf{p}}_2, \dots, \hat{\mathbf{p}}_{S}$ corresponding to the $S$ temporal branches.

\textbf{Tokenizer decoder.} The purpose of the decoder module (a series of shared transformer layers followed by prediction heads) is to reconstruct the original signal information based on the learned codebook tokens. Directly reconstructing raw biosignals has been shown to yield unstable training and poor performance, due to the inherently noisy and high-dimensional nature of these signals. Therefore, we leverage the Fourier spectrum (more specifically the phase $\phi$ and spectral amplitude $A$) as a more structured target for reconstruction.

\textbf{Modality-adaptive design parameters.} \sys tokenizer architecture exposes four principled parameters that can be tuned per modality to accommodate biosignals with fundamentally different spectro-temporal characteristics: (\emph{i}) the patch length $w$, which controls the temporal context per patch; (\emph{ii}) the number of temporal branches $S$, which determines how many frequency scales the signal is decomposed into; (\emph{iii}) the temporal convolution kernel sizes $K_{\text{temporal}_1}, \dots, K_{\text{temporal}_{S}}$, which align each branch's receptive field with the characteristic time scales of the target modality; and (\emph{iv}) the number of RVQ codebooks N per branch, which dictates the quantization depth.

\subsection{Phase-Aware Training Loss}
\label{sec:tokenizer_training_analysis}


However, architecture alone does not account for \sys's reconstruction gains: the training loss is equally critical. The \sys tokenizer loss operates in the Fourier domain (phase and amplitude) and consists of three complementary loss components.

\paragraph{Phase loss.} Reconstructing the phase $\phi$ directly with mean squared error (MSE) is suboptimal due to the periodic nature of the Fourier phase (see Appendix \ref{apx:sine/cosine}). To address this, the tokenizer decoder replaces raw phase reconstruction with a sine/cosine phase representation and predicts two values $\hat{c}$ and $\hat{s}$, corresponding to estimates of $\cos\phi$ and $\sin\phi$ respectively, which are smooth, continuous and better suited as loss terms. Since both predictions $\hat{c}$ and $\hat{s}$ refer to the same phase angle, we optimize them jointly rather than independently. We represent each phase as a 2D vector and compute the cosine similarity between the predicted vector ($\hat{c}$, $\hat{s}$) and the target ($\cos \phi, \sin \phi$). This formulation naturally respects the circular topology of phase: the phases are represented as vectors in 2D space, making the errors between them geometrically meaningful. 

\paragraph{Amplitude loss.} For the amplitude component, rather than reconstructing raw spectral magnitudes $A$, we operate in the log-amplitude domain $\log(1 + A)$. This logarithmic scaling compresses the large dynamic range of signal spectra and places relatively greater emphasis on reconstructing high-frequency components. Therefore, the tokenizer decoder has three prediction heads responsible for reconstructing $\log(1+\hat{A})$, $\hat{c}$ and $\hat{s}$, respectively.



\paragraph{Temporal loss.} Finally, although reconstructing raw biosignal $X$ waveforms might lead to instability issues \citep{LaBraM}, we found that incorporating a temporal-domain reconstruction loss provides additional guidance. Specifically, we transform the predicted log-amplitude and phase back to the time domain (the predicted log-amplitudes are exponentiated and rescaled to recover the original amplitude spectrum before the inverse Fourier transformation) and compute a temporal MSE regularization term. 

The \sys tokenizer is thus trained end-to-end with the objective:


\vspace{-0.1in}
\begin{equation}
\begin{aligned}
\mathcal{L}_T
&= \sum_{i}
\underbrace{
\left\| \log (1+\hat{A_i}) - \log (1+A_i) \right\|_2^2
}_{\textcolor{purple}{\text{Log-Amplitude Loss}}} +\;
\underbrace{
1-\sum_{i}
\frac{\hat{c}_i \cos{\phi_i} + \hat{s}_i \sin{\phi_i}}
{\sqrt{\hat{c}_i^2 + \hat{s}_i^2}}
}_{\textcolor{blue}{\text{Phase Loss}}}
\;
+\;
\underbrace{
\left\| \hat{X_i} - X_i \right\|_2^2
}_{\textcolor{orange}{\text{Temporal Loss}}}
+\;
\mathcal{L}_Q
\end{aligned}
\label{eq:labram_plus_loss}
\end{equation}

where $\hat{A}$, $\hat{c}$, and $\hat{s}$ denote reconstructed Fourier components, $\hat{X}$ denotes reconstructed biosignal in the time domain and $\mathcal{L}_Q$ is the commitment loss.

As shown in Appendix~\ref{apx:unit-circle}, the phase-aware loss is a critical component. The principled combination of multi-scale architecture (Section \ref{sec:model_arch_tokenizer}) and its novel training loss (Section \ref{sec:tokenizer_training_analysis}) is what enables \sys's high-fidelity reconstruction. 

\section{Modality-Specific Configuration}
\label{sec:modality_config}

The \sys tokenizer architecture is modality-agnostic by design, but its parameters (namely patch length, sampling rate, number of temporal branches and RVQ codebooks) must be adapted to the distinct spectro-temporal characteristics of each biosignal. In this section, we describe the signal properties and the scaling analysis that motivate our configuration choices.

\subsection{Signal Characteristics}
\label{sec:signal_resolution}

\paragraph{EEG.} 
Neural oscillations of interest span distinct frequency bands, namely delta (0.5--4\,Hz), theta (4--8\,Hz), alpha (8--13\,Hz), beta (13--30\,Hz), and gamma ({>}\,30\,Hz). Each band is linked to specific cognitive states, with clinically and cognitively relevant content concentrated below 100\,Hz ~\citep{Newson2018_lm}. We resample all EEG recordings to 200\,Hz, which satisfies the Nyquist criterion for frequencies up to 100\,Hz. We set the patch length to $w=1sec$, offering a practical trade-off between temporal localization and spectral resolution for downstream tokenization.

\paragraph{ECG.} The cardiac electrical cycle is characterized by a sequence of morphologically distinct waveforms whose shapes and duration carry critical diagnostic information: the P wave ($\sim60-100\,ms$), the QRS complex ($\sim80-120\,ms$) and the T wave (typically $\sim160\,ms$) ~\citep{Sattar2023}. The clinically relevant frequency content of ECG signals is largely concentrated below 100\,Hz, with high-frequency components above this range primarily reflecting noise ~\citep{Kligfield2007}. Accordingly, we resample ECG recordings to 200\,Hz. We set the patch length to $w=200ms$, a window that is sufficiently long to capture the QRS complex while remaining short enough to isolate individual waveform components without mixing adjacent cardiac features. 

\paragraph{EMG.} EMG signals arise from the superposition of Motor Unit Action Potentials (MUAPs), which are brief electrical discharges lasting approximately 5--30\,ms ~\citep{Dumitru1997}. The resulting surface EMG signal contains significant spectral energy extending up to 500\,Hz, with the dominant power typically concentrated between 20 and 250\,Hz ~\citep{deluca1997use}. To adequately capture these high-frequency dynamics, we resample EMG recordings to 1000\,Hz (substantially higher than for EEG and ECG). We set the patch length to $w=200ms$, which is long enough to encompass multiple MUAP events while remaining short enough to track rapid changes in muscle activation.


Although \sys design is flexible, these modality-specific signal-level choices also influence the design of the temporal encoder, particularly the choice of kernel sizes (see Appendix \ref{apx:hypers_tokenizer}). 

\subsection{Scaling Analysis}
\label{sec:scaling_analysis}


\begin{figure}[!h]
    \centering
\includegraphics[width=\linewidth]{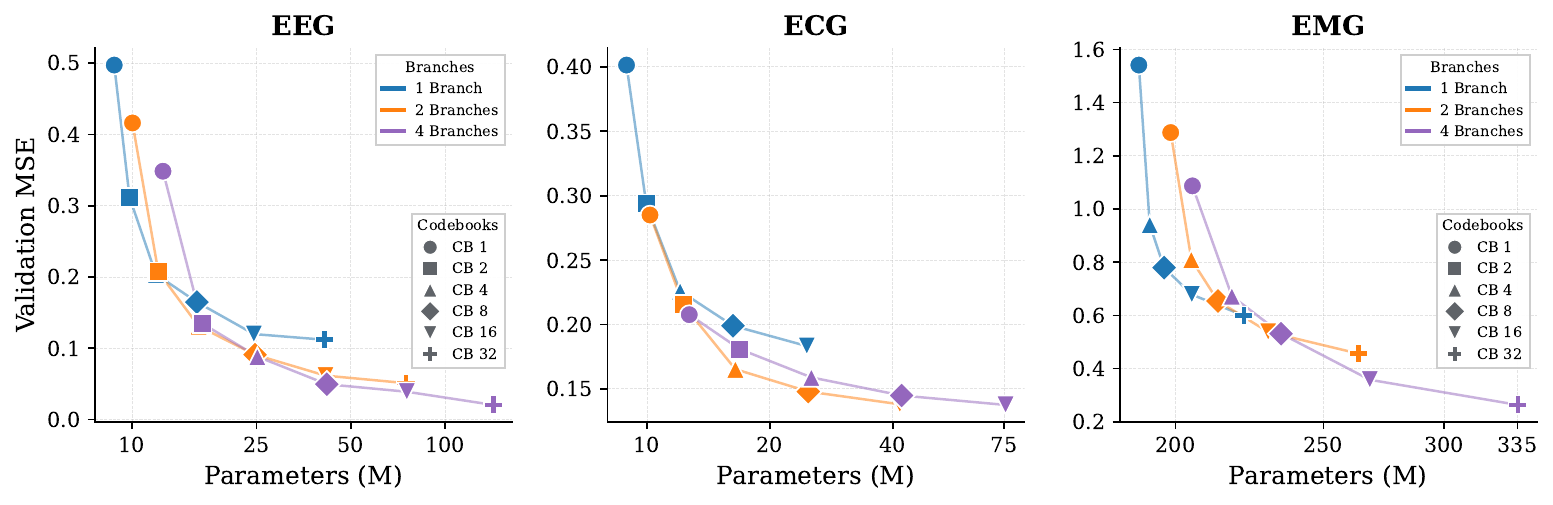}
    \caption{Scaling Analysis. Mean squared error (MSE) reconstruction performance and number of trainable parameters for \sys with various numbers of temporal branches and RVQ codebooks (CB) per branch across three modalities (EEG, ECG, EMG).}
\label{fig:scaling_laws}
\end{figure}

To determine the optimal tokenizer configuration for each modality, we conduct a systematic scaling analysis over two key architectural hyperparameters \footnote{Larger configurations were not explored due to computational budget constraints.}: (\emph{i}) the number of temporal branches $S \in \{1,2,4\}$ and (\emph{ii}) the number of RVQ codebooks per branch $N \in \{1,2,4,8,16,32\}$. For each modality, we select a compact representative dataset, namely PhysioNet MI~\citep{physionetmi} for EEG, PTB~\citep{Bousseljot2004-wi} for ECG, and the small version of emg2pose~\citep{salteremg2pose} for EMG. We train the \sys tokenizer for 50 epochs using the settings described in Appendix~\ref{apx:hypers_tokenizer}. We report validation MSE (in an unseen set) averaged over the last 10  epochs. 

\paragraph{General trends.} As seen from the results in Figure~\ref{fig:scaling_laws} (separate plots for parameter counts and reconstruction MSE as a function of codebook depth / number of branches are provided in Appendix~\ref{apx:scaling_details}), two consistent patterns emerge across all modalities: (\emph{i}) increasing the number of codebooks per branch yields large reductions in reconstruction error, confirming that deeper residual quantization refines the discrete approximation of continuous signal features. (\emph{ii}) adding temporal branches provides complementary gains (particularly when the number of codebooks is small) by adding specific priors and enabling the model to decompose the signal into frequency-specific representations that are each easier to quantize. When both dimensions are scaled jointly, the improvements accumulate.

\paragraph{Modality-specific observations.} 
For modalities sampled at the same rate (200Hz), signal complexity dictates the saturation behavior: less noisy ECG saturates earlier, whereas EEG with richer spectral structure demands finer-grained quantization, continues to benefit from scaling the number of codebooks. For EMG, which is sampled at a substantially higher rate (1000Hz) to capture its broadband spectral content, the steepest improvements come from increasing codebook depth, suggesting that the wider frequency range requires greater representational capacity (codebook utilization analysis is provided in Appendix \ref{apx:codebook_utilization}). These trends validate a core design principle of \sys: the multi-scale branch structure and hierarchical RVQ codebooks are complementary mechanisms whose optimal balance is modality-dependent. 

\section{Evaluation}

\subsection{Signal Reconstruction Evaluation}
\label{sec:setup_tokenizer}

\paragraph{Tokenizer training.} For each biosignal modality, a \sys tokenizer was trained using the hyperparameters described in Appendix~\ref{apx:hypers_tokenizer} on the datasets listed in Appendix~\ref{apx:dataset}. All datasets were resampled to the modality-specific sampling frequency defined in Section~\ref{sec:signal_resolution}. Based on the scaling analysis in Section~\ref{sec:scaling_analysis} and available computational resources, we select $S=4$ temporal branches for all modalities with $N=8$ codebooks per branch for EEG and ECG and $N=16$ for EMG, reflecting the greater representational demands of broadband EMG signals. All experiments were conducted on an NVIDIA DGX system with 4 Tesla V100 GPUs. 

\begin{figure}[!h]
  \centering
  \includegraphics[width=\linewidth]{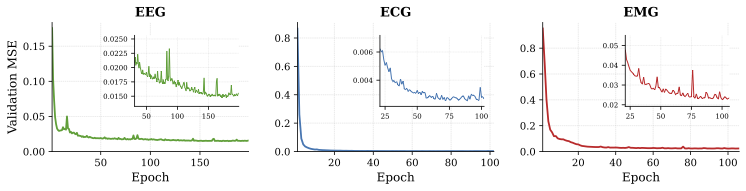}
  \caption{
  Validation mean squared error (MSE) during \sys tokenizer training across all three modalities. Note that absolute MSE values are not directly comparable across modalities.}
  \label{fig:val_curves_all}
\end{figure}

\paragraph{Reconstruction results.} 

As shown in Figure \ref{fig:val_curves_all}, \sys achieves \emph{high-fidelity signal reconstruction} across all modalities. ECG exhibits the fastest and sharpest convergence consistent with its stereotyped morphology, EEG converges to a stable plateau reflecting its richer spectral structure and EMG converges more gradually, consistent with its broadband nature. For ECG and EMG, no unimodal codebook-based tokenizers currently exist; \sys is the first to demonstrate high-fidelity codebook-based reconstruction for these modalities (reconstruction examples are provided in Appendix~\ref{apx:validation_curves}). For EEG, we compare \sys against two pretrained open-source codebook-based tokenizers: LaBraM
and BrainOmni.
\sys consistently outperforms both across evaluation setups (see Appendix~\ref{apx: other_tokenizers} for detailed comparisons).

\subsection{Downstream Task Performance}
\label{sec:downstream}

To demonstrate that high-fidelity tokenization translates into strong downstream performance, we evaluate fine-tuned \sys foundation models (\sys-FMs) across all three biosignal modalities using the tokenizers of Section \ref{sec:setup_tokenizer}.

\begin{figure}[!h]
    \centering
    \includegraphics[width=\linewidth]{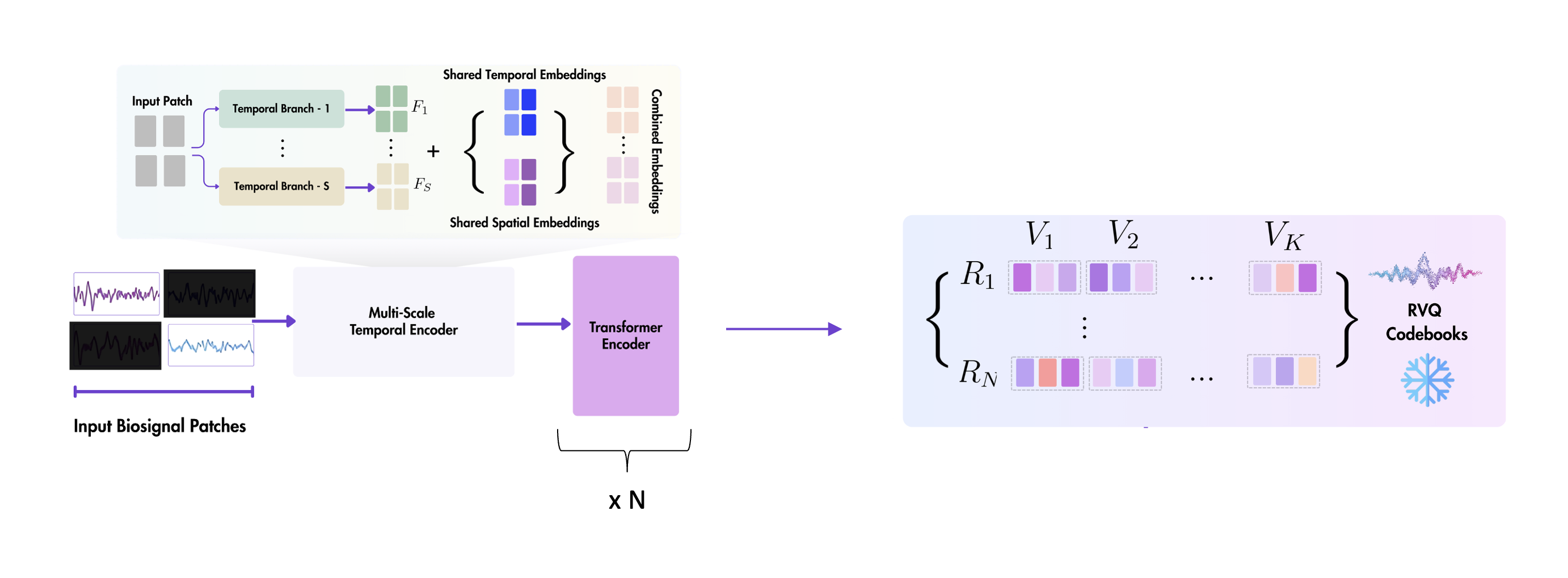}
     \caption{{\small The {\sffamily\neurorvq}-FM model. The multi-scale temporal encoder extracts multi-scale temporal features that are passed to a series of transformer encoder layers. The training objective is to reconstruct masked tokens using the learned codebooks $\mathcal{R}$}.}
    \label{fig:neurorvq_fm}
\end{figure}

\paragraph{Foundation model architecture and training.} \sys-FM operates on the tokenized representation produced by the \sys tokenizer. The architecture consists of the multi-scale temporal encoder (Section~\ref{sec:model_arch_tokenizer}), followed by transformer layers that model long-range dependencies and token prediction heads that reconstruct masked tokens using the learned codebooks $\mathcal{R}$ (see Figure \ref{fig:neurorvq_fm}). The model is pre-trained with a masked-token prediction objective using a symmetric masking strategy~\citep{LaBraM}. For downstream tasks, the pre-trained backbone is extended with task-specific classification heads. We deliberately keep the foundation model architecture simple to isolate the contribution of the tokenizer: improvements in downstream performance can be attributed to the quality of the learned token representations rather than architectural complexity. The hyperparameters for the pre-training and fine-tuning of \sys foundation model for each biosignal modality are described in Appendix~\ref{apx:hypers_foundation} on the datasets listed in Appendix~\ref{apx:dataset}. All datasets were resampled to the modality-specific sampling frequency defined in Section~\ref{sec:signal_resolution}.

\paragraph{Downstream task evaluation.} We evaluate fine-tuned \sys-FMs across all three modalities. For EEG, we compare against nine foundation models on five downstream classification tasks from the benchmark of~\citet{lee2025assessing}. These models include two models that use codebook-based tokenizers like \sys, namely LaBraM and BrainOmni and seven common foundation model baselines, namely BIOT\footnote{BIOT could not be tested on Sleep since the benchmark electrodes are missing from its pre-trained model. Its pre-trained bipolar channels were mapped to standard locations.}, MIRepNet\footnote{As a motor-specific foundation model, MIRepNet was tested only on the Motor task.}, CBraMod, EEGPT, NeuroGPT, 
CSBrain and REVE. For ECG, we compare against two foundation models, HuBERT-ECG and ECGFounder, on the PTB-XL benchmark~\citep{PTB-XL-Nature}. For EMG, we compare against two foundation models, PhysioWave and TinyMyo, on four classification datasets (Discrete Gestures~\citep{Kaifosh2025-fi}, EPN-612~\citep{benalcazarEMGEPN612Dataset2020}, NinaPro DB5~\citep{pizzolatoComparisonSixElectromyography2017}, UCI-EMG~\citep{emg_data_for_gestures_481}). Appendix~\ref{apx:downstream_finetuning_analysis} provides fine-tuning, implementational details and further analysis. Appendix~\ref{apx:frozen_head_section} presents the head-only fine-tuning results for all \sys-FMs and other foundation model baselines.

\begin{table*}[!t]
  \caption{Downstream task performance across three biosignal modalities. \textbf{Bold} and \underline{underlined} values indicate best and second-best performance per column, respectively. All evaluations are subject-independent. EEG: Balanced accuracy (BAcc). ECG: Accuracy and Balanced accuracy (BAcc). EMG: Accuracy, F1, Balanced Accuracy (BAcc) and CLER (Classification Error Rate).}
  \label{table:downstream_all}
  \centering

\vspace{0.3em}
  {\small\textsc{(a) EEG --- Five Classification Tasks~\citep{lee2025assessing}}}\\[0.4em]
  \resizebox{0.75\textwidth}{!}{
  \begin{tabular}{l|ccccc}
    \toprule
    Model (Backbone Params) & Motor & ERP & Memory & Sleep & Eyes \\
    \midrule
    BIOT~\citep{NEURIPS2023_f6b30f3e} (3.2M) & 44.3{\scriptsize$\pm$7.9} & 50.0{\scriptsize$\pm$0.0} & 51.0{\scriptsize$\pm$1.8} & -- & 76.3{\scriptsize$\pm$4.9} \\
    MIRepNet~\citep{liu2025mirepnetpipelinefoundationmodel} (5.2M) & 68.9{\scriptsize$\pm$8.6} & -- & -- & -- & -- \\
    \midrule
    CBraMod~\citep{wang2025cbramod} (4.9M)  & 61.4{\scriptsize$\pm$10.4} & 77.7{\scriptsize$\pm$5.2} & 57.4{\scriptsize$\pm$3.8} & 63.5{\scriptsize$\pm$4.1} & 83.9{\scriptsize$\pm$4.1}\\
    
    EEGPT~\citep{wang2024eegpt}  (25.7M)   & 31.3{\scriptsize$\pm$3.5} & 66.8{\scriptsize$\pm$14.6} & 52.0{\scriptsize$\pm$1.7} & 63.4{\scriptsize$\pm$4.4} & 79.7{\scriptsize$\pm$3.7} \\
    
    NeuroGPT~\citep{NeuroGPT} (79.5M)  & 68.2{\scriptsize$\pm$8.3} & 75.7{\scriptsize$\pm$4.8} & \textbf{59.7{\scriptsize$\pm$2.9}} & \underline{67.4{\scriptsize$\pm$3.3}} & 82.7{\scriptsize$\pm$3.6} \\

    CSBrain~\citep{zhou2025csbrain} (4.9M) & {67.1\scriptsize$\pm$11.1} & {76.5 \scriptsize$\pm$5.7} & {57.0\scriptsize$\pm$3.5} & {67.2\scriptsize$\pm$3.8} & {81.9\scriptsize$\pm$4.0}  \\

    REVE~\citep{ouahidi2025reve} (69.2M) & \underline{{69.4\scriptsize$\pm$10.5}} & \underline{{82.6\scriptsize$\pm$4.3}} & \underline{{59.3\scriptsize$\pm$3.2}} & {62.9\scriptsize$\pm$3.5} & {79.9\scriptsize$\pm$3.4} \\
    
    \midrule
    BrainOmni$^{\dagger}$ ~\citep{xiao2025brainomni} (37.8M) & 58.5{\scriptsize$\pm$5.8} & 72.3{\scriptsize$\pm$4.1} & 51.8{\scriptsize$\pm$2.3} & 57.6{\scriptsize$\pm$5.5} & \underline{85.2{\scriptsize$\pm$2.2}} \\
    
    LaBraM$^{\dagger}$ ~\citep{LaBraM} (5.8M) & 63.0{\scriptsize$\pm$7.6} & 82.2{\scriptsize$\pm$4.0} & 52.6{\scriptsize$\pm$2.6} & 65.2{\scriptsize$\pm$3.7} & 79.9{\scriptsize$\pm$4.7} \\
    \midrule
    
    \sys-EEG$^{\dagger}$ (5.9M)  & \textbf{70.0{\scriptsize$\pm$7.3}} & \textbf{87.6{\scriptsize$\pm$3.3}} & 57.4{\scriptsize$\pm$2.7} & \textbf{72.8{\scriptsize$\pm$2.8}} & \textbf{86.9{\scriptsize$\pm$2.6}} \\
    \bottomrule
  \end{tabular}
  }\\[0.2em]
  \label{tab:downstream_results_all_modalities}
  {\scriptsize $^{\dagger}$\, Models that use codebook-based tokenizers}

  \vspace{0.5em}

  {\small\textsc{(b) ECG --- PTB-XL Classification}}\\[0.4em]
  \tiny
  \resizebox{0.60\textwidth}{!}{
  \begin{tabular}{l|cc|cc}
    \toprule
    & \multicolumn{2}{c|}{\textbf{5-class}} & \multicolumn{2}{c}{\textbf{43-class}} \\
    Model (Backbone Params) & Accuracy & BACC & Accuracy & BACC \\
    \midrule
    HuBERT-ECG ~\citep{HuBERT} (30.2M) & \underline{72.60} & 60.23 & 62.49 & 20.71 \\
    ECGFounder ~\citep{ECGFounder} (30.7M) & \textbf{76.55} & \textbf{65.39} & \underline{65.51} & \underline{28.96} \\
    \midrule
    \sys-ECG (0.3M) & 70.19 & \underline{64.50} & \textbf{79.17} & \textbf{58.33} \\
    \bottomrule
  \end{tabular}
  }

  \vspace{1.0em}

  {\small\textsc{(c) EMG --- Four Classification Tasks}}\\[0.4em]
  \resizebox{\textwidth}{!}{
  \begin{tabular}{l|cc|cc|cc|cc}
    \toprule
    & \multicolumn{2}{c|}{\textbf{Discrete Gestures}} & \multicolumn{2}{c|}{\textbf{EPN-612}} & \multicolumn{2}{c|}{\textbf{NinaPro DB5}} & \multicolumn{2}{c}{\textbf{UCI-EMG}} \\
    Model (Backbone Params) & BAcc\,$\uparrow$ & CLER\,$\downarrow$ & Acc & F1 & Acc & F1 & Acc & F1 \\
    \midrule
    PhysioWave ~\citep{chen2025physiowave} (4.8M) & 54.70{\scriptsize$\pm$0.0} & 64.20{\scriptsize$\pm$0.9} & 90.30{\scriptsize$\pm$0.6} & 90.35{\scriptsize$\pm$0.3} & 24.91{\scriptsize$\pm$2.5} & 22.95{\scriptsize$\pm$2.5} & 56.52{\scriptsize$\pm$2.4} & 55.76{\scriptsize$\pm$2.8} \\
    TinyMyo ~\citep{fasulo2026tinymyotinyfoundationmodel} (3.6M)  & 39.70 & 64.20 & 84.68{\scriptsize$\pm$1.9} & 84.68{\scriptsize$\pm$1.9} & 25.26{\scriptsize$\pm$4.5} & 23.29{\scriptsize$\pm$4.9} & \underline{85.99{\scriptsize$\pm$0.4}} & \underline{85.66{\scriptsize$\pm$0.4}} \\
    \midrule
    \sys-EMG (5.9M) & \textbf{70.80{\scriptsize$\pm$0.6}} & \textbf{27.60{\scriptsize$\pm$0.8}} & \textbf{94.65{\scriptsize$\pm$0.4}} & \textbf{94.66{\scriptsize$\pm$0.5}} & \textbf{41.36{\scriptsize$\pm$2.8}} & \textbf{38.76{\scriptsize$\pm$3.7}} & \textbf{89.43{\scriptsize$\pm$1.1}} & \textbf{89.28{\scriptsize$\pm$1.1}} \\
    \bottomrule
  \end{tabular}
  }
\end{table*}

As shown in Table~\ref{table:downstream_all}, \sys achieves strong downstream performance across all three biosignal modalities. \sys-EEG achieves the best balanced accuracy on four of five tasks and the highest mean balanced accuracy overall. \sys-ECG achieves high balanced accuracy on both PTB-XL settings, doubling the balanced accuracy of the next-best model on the fine-grained 43-class task. 
ECGFounder leads on the 5-class setting with \sys in the second-best bacc while in the 43-class setting \sys outperforms all baselines. \sys-EMG outperforms PhysioWave and TinyMyo across all four classification tasks by a wide margin. Across all three modalities, \sys's consistent improvements with a compact and simple foundation model architecture confirm the central claim of this work: \emph{high-fidelity tokenization is the critical ingredient for effective biosignal foundation models} and its benefits transfer across physiologically diverse signal types. To isolate the effect of high-fidelity tokenization in downstream task performance, we conducted a tokenizer-swap ablation study (Appendix \ref{apx:swap_ablation}). The results show that tokenization quality translates into downstream performance: balanced accuracy improves with tokenizer fidelity and the effect is clearest under full fine-tuning.

\section{Discussion}

The \sys  tokenizer demonstrates state-of-the-art reconstruction capabilities (Section \ref{sec:setup_tokenizer}) on out-of-distribution and in-distribution signals (Appendix \ref{apx: other_tokenizers} and Appendix \ref{apx:in_distribution_curves}). Importantly, these results cannot be attributed solely to the number of codebooks (single or multiple), their size or their type (e.g., RVQ-based). Instead, they arise from the inductive biases embedded in the tokenizer design (Section \ref{sec:modality_config}) which, together with the newly introduced phase-aware training loss (Section \ref{sec:tokenizer_training_analysis} and Appendix \ref{apx:unit-circle}), are particularly well-suited for capturing the complex nature of biosignals. 

Since biosignals differ fundamentally in their spectro-temporal characteristics, we train separate tokenizers in this work, each optimally configured for its target modality while the architecture remains identical: only signal-level hyperparameters change (see Section~\ref{sec:modality_config}). Whether a shared cross-modal tokenizer can match this per-modality performance is an interesting future direction.

A natural concern is whether pursuing high-fidelity reconstruction is warranted for inherently noisy signals. However, high-fidelity reconstruction does not mean memorizing noise. It ensures sufficient representational capacity to preserve all signal components, including subtle discriminative features that overlap spectrally with noise, letting downstream foundation models learn which are task-relevant. 

The main contribution of this work is the \sys tokenizer, which can be integrated into any biosignal foundation model. Further improvements can come from larger foundation architectures, cross-modal pre-training and incorporating causal reasoning-inspired training principles such as more targeted temporal and spatial masking strategies~\citep{barmpas2024a}. 


\section{Conclusion}

In this work, we introduce \sys, a modality-adaptive biosignal tokenizer family designed for high-fidelity signal reconstruction across EEG, ECG and EMG. Our main contribution is the design and training of the tokenizer, which captures low- and high-frequency components of biosignals, achieves state-of-the-art reconstruction performance and facilitates the development of more effective biosignal foundation models. We showed that when a simple masked-token foundation model is trained for each modality using  \sys tokenizers, it outperforms existing pre-trained open-source modality-specific foundation models in various downstream tasks. In summary, \sys advances the development of efficient large  biosignal foundation models by providing a high-fidelity biosignal tokenizer, paving the way for broader use of foundation models in the biosignal space.

\section*{Acknowledgments}

K. Barmpas was responsible for the development of NeuroRVQ, including dataset preprocessing, tokenizer analysis and models' training. K. Barmpas and S. Zafeiriou jointly designed NeuroRVQ's theoretical formulation. N. Lee conducted the EEG and ECG benchmarking analysis. D. Chalatsis and W. Raftery conducted the EMG benchmarking analysis.  Y. Panagakis, D. Adamos, N. Laskaris, A. Koliousis and D. Farina provided valuable guidance and feedback throughout the project.

This work was supported by the EPSRC Turing AI Fellowship (Grant Ref: EP/Z534699/1), \textit{Generative Machine Learning Models for Data of Arbitrary Underlying Geometry (MAGAL)}.

\newpage
\bibliographystyle{plainnat}
\bibliography{example_paper}

\newpage
\appendix
\onecolumn
\section{Design of the Phase Term in the Phase-Aware Loss}
\label{apx:sine/cosine}

We describe the benefits of using a sine-cosine representation of the Fourier phase $\phi$ in the loss function, instead of the raw phase values.

Other tokenizer architectures (e.g LaBraM~\citep{LaBraM}) compute phase loss directly as the squared error between predicted $\hat{\phi}$ and true $\phi$: 
\begin{equation}
    \mathcal{L}_\phi = ||\hat{\phi}-\phi||^{2}.
\end{equation}
This formulation, however, is discontinuous at the boundaries of the phase domain $[-\pi, \pi]$. For instance, with $\hat{\phi} = \pi - \epsilon$ and $\phi = -\pi + \epsilon$ for small $\epsilon > 0$,
\begin{equation}
    \mathcal{L}_\phi = ||\hat{\phi}-\phi||^{2} = (2\pi - 2\epsilon)^{2}.
\end{equation}
The loss approaches $4\pi^{2}$, even though the two angles are nearly identical.
Thus, small phase shifts near the $\pm \pi$ boundary (e.g., $2^\circ$ between $179^\circ$ and $-179^\circ$) result in large discontinuous jumps in the loss,
leading to unstable gradients and poor convergence.

To address this issue, we map phase $\phi$ to its sine and cosine components, a point on the unit circle:
\begin{equation}
    r(\phi) = (\sin(\phi), \cos(\phi)) \in \mathbb{R}^2,
\end{equation}
The corresponding loss is: 
\begin{equation}
    \mathcal{L}_{r(\phi)} = ||\sin(\hat{\phi}) - \sin(\phi)||^2 + ||\cos(\hat{\phi}) - \cos(\phi)||^2 = 2 - 2\cos(\hat{\phi} - \phi)
\end{equation}
Our formulation is continuous across the $\pm \pi$ boundary.
Revisiting the earlier example with $\hat{\phi} = \pi - \epsilon$ and $\phi = -\pi + \epsilon$, 
\begin{equation}
\mathcal{L}_{r(\phi)} = 2 - 2\cos(2\epsilon) \approx 4\epsilon^2
\end{equation}
which smoothly approaches $0$ as $\epsilon \rightarrow 0$. 


While $\mathcal{L}_{r(\phi)}$ resolves the discontinuity problem, it creates a gradient issue when the prediction is maximally wrong on the unit circle, i.e $\hat{\phi} \approx \phi \pm \pi$. Consider the gradient with respect to $\hat{\phi}$:
\begin{equation}
    \frac{\partial}{\partial \hat{\phi}} \mathcal{L}_{r(\phi)} = 2\sin(\hat{\phi} - \phi)
\end{equation}

When $\hat{\phi} \approx \phi \pm \pi$, $\sin(\hat{\phi} - \phi) \approx 0$. This means the worst predictions receive weak learning update.

Instead, our architecture predicts two values $\hat{c}$ and $\hat{s}$ (see Section \ref{sec:tokenizer_training_analysis}), corresponding to estimates of 
$\cos\phi$ and $\sin\phi$ respectively. 
To avoid the issues relating to $\mathcal{L}_{r(\phi)}$ but keeping its benefits, we introduce a joint optimization of $\hat{c}$ and $\hat{s}$ with the phase-aware loss term $\mathcal{L}_{\text{phase}}$, since both values refer to the same underlying phase angle. $\mathcal{L}_{\text{phase}}$ optimizes the 
directional alignment between the predicted vector $(\hat{c}, \hat{s})$ 
and the target $(\cos\phi, \sin\phi)$ via cosine similarity. Since the 
target is a unit vector by definition ($\cos^2\phi + \sin^2\phi = 1$), 
the loss simplifies to:
\begin{equation}
    \mathcal{L}_{\text{phase}} = 1 - \sum_{i} 
    \frac{\hat{c}_i \cos\phi_i + \hat{s}_i \sin\phi_i}
    {\sqrt{\hat{c}_i^2 + \hat{s}_i^2}}
\end{equation}
The gradient of $\mathcal{L}_{\text{phase}}$ with respect to the predicted 
vector is well-behaved across the full range of phase errors: it is 
proportional to the component of the predicted vector orthogonal to the 
target direction, scaled by the inverse of the predicted norm. This means 
that (\emph{i}) maximally wrong predictions produce large gradients and (\emph{ii}) the loss jointly optimizes 
direction rather than treating $\hat{c}$ and $\hat{s}$ as independent quantities. 

\begin{figure}[!h]
    \centering
    \begin{subfigure}[t]{0.45\linewidth}
        \centering
        \includegraphics[width=0.7\linewidth]{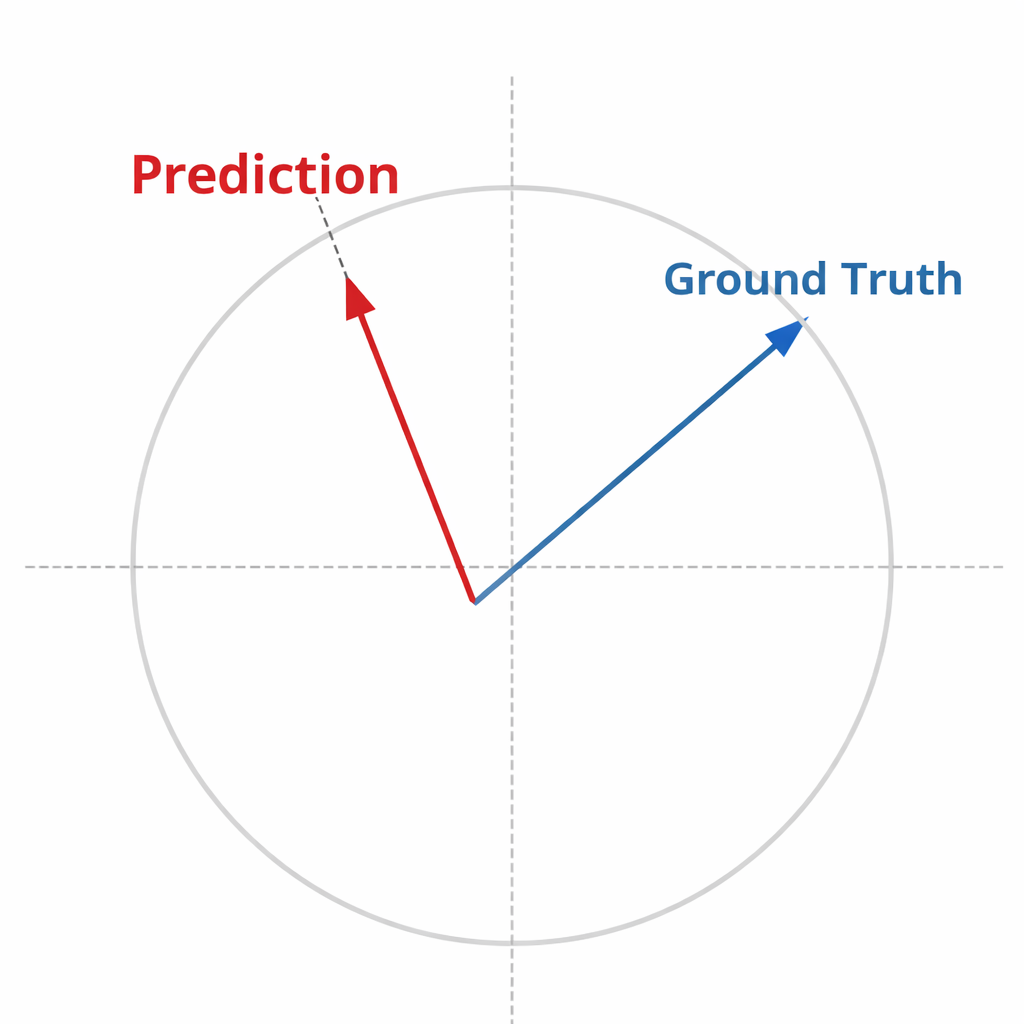}
        \caption{Prediction  off the unit circle.}
        \label{fig:mse}
    \end{subfigure}
    \hfill
    \begin{subfigure}[t]{0.45\linewidth}
        \centering
        \includegraphics[width=0.7\linewidth]{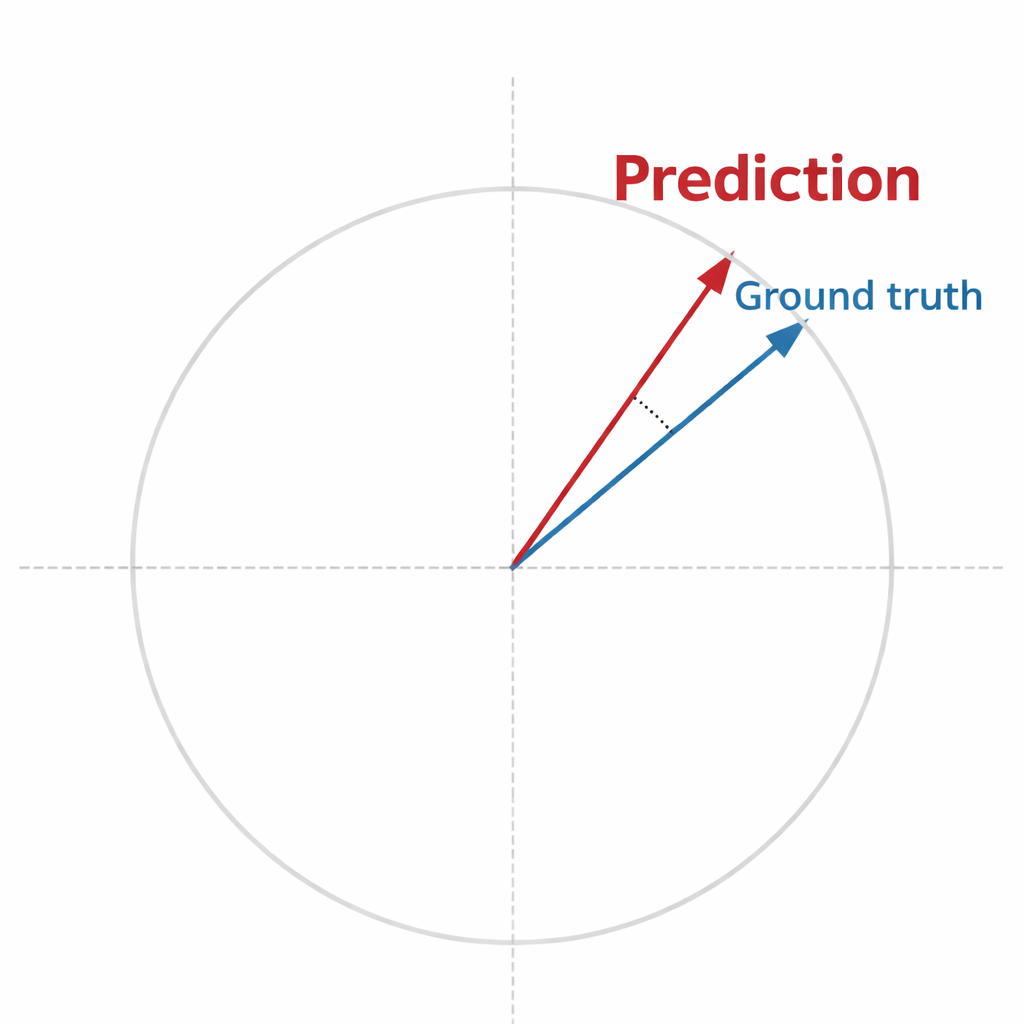}
        \caption{Prediction on the unit circle.}
        \label{fig:unitcircle}
    \end{subfigure}
    \caption{Visualization of phase predictions off (a) and on (b) the unit circle.}
    \label{fig:phase_comparison}
\end{figure}

Although cosine 
similarity is scale-invariant and does not explicitly constrain $(\hat{c}, 
\hat{s})$ to lie on the unit circle (see Figure \ref{fig:phase_comparison}), the predicted norms tend to remain 
close to unity in practice. Let $r = \sqrt{\hat{c}^2 + \hat{s}^2}$ denote the predicted norm. The gradient of $\mathcal{L}_{\text{phase}}$ with respect to 
$\hat{c}$ is:
\begin{equation}
    \frac{\partial \mathcal{L}_{\text{phase}}}{\partial \hat{c}} = 
    -\frac{1}{r}\left(\cos\phi - \frac{\hat{c}\cos\phi + \hat{s}\sin\phi}{r^2}\hat{c}\right)
\end{equation}

The $1/r$ scaling optimizes direction without creating incentives to grow or shrink the predicted norm. In practice, predicted norms remain close to unity 
throughout training. This is primarily because the decoder's $\tanh$ output activation bounds $\hat{c}$ and $\hat{s}$ to $[-1, 1]$, constraining 
$r \leq \sqrt{2}$ and preventing norm divergence. As the predictions converge toward the targets $(\cos\phi, \sin\phi)$, the norm naturally 
approaches 1. Nonetheless, an optional regularization term can be added to explicitly penalize deviations from unit circle:
\begin{equation}
    \mathcal{L}_{\text{reg}} = \lambda_{\text{circle}} \cdot 
    \sum_{i} \left( \hat{c}_i^2 + \hat{s}_i^2 - 1 \right)^2
\end{equation}

However, this introduces a trade-off: the regularization gradient is 
radial (adjusting magnitude), while the cosine similarity gradient is 
tangential (adjusting direction). Any nonzero $\lambda_{\text{circle}}$ diverts part of each optimization step toward magnitude correction rather than directional alignment, slightly reducing the efficiency of phase learning. Our sensitivity analysis  (Appendix~\ref{apx:unit-circle}) confirms this: 
$\lambda_{\text{circle}} = 0.0$ achieves the lowest reconstruction 
MSE, with performance degrading as $\lambda_{\text{circle}}$ increases. In our main results, however, we retain $\lambda_{\text{circle}} = 0.1$ as a conservative choice that provides an explicit geometric guarantee at negligible cost, ensuring that predicted phase vectors remain valid across the diverse signal distributions encountered when training on  multiple datasets and modalities. In summary, our approach combines the strengths of all formulations while avoiding their respective weaknesses:
\begin{center}
\small
\begin{tabular}{lcccc}
    \toprule
    & \shortstack{Continuous\\at $\pm\pi$} & \shortstack{No vanishing\\gradients} & \shortstack{Joint\\optimization} & \shortstack{Unit-circle} \\
    \midrule
    $\mathcal{L}_\phi$ & \ding{55} & \ding{51} & -- & -- \\
    $\mathcal{L}_{r(\phi)}$ & \ding{51} & \ding{55} & \ding{55} & \ding{51} \\
    $\mathcal{L}_{\text{phase}}$ & \ding{51} & \ding{51} & \ding{51} & $\approx$\textsuperscript{*} \\
    $\mathcal{L}_{\text{phase}}$\ + $\mathcal{L}_{\text{reg}}$ & \ding{51} & \ding{51} & \ding{51} & \ding{51}\textsuperscript{*} \\
    \bottomrule
\end{tabular}\\[0.3em]
{\scriptsize \textsuperscript{*}Approximate unit-circle enforcement via $\tanh$ activation. \\ $\mathcal{L}_{\text{reg}}$ provides exact enforcement but introduces a convergence trade-off (Appendix~\ref{apx:unit-circle}).}

\end{center}

\section{Architectural and Training Components Analysis}
\label{apx:unit-circle}

\subsection{Individual Component Contributions}

As described in Section~\ref{sec:tokenizer_training_analysis}, the \sys training objective consists of multiple components. To quantify the individual contribution of each architectural and training component, we conduct the following ablation studies using the datasets described in Section \ref{sec:scaling_analysis}. We start from a simple baseline: a single-branch, single-codebook architecture that naively reconstructs the Fourier amplitude and phase using MSE (as discussed in Appendix~\ref{apx:sine/cosine}). We progressively introduce to this baseline each \sys training objective component. 

\begin{table}[!h]
  \caption{Individual training component contribution analysis. Starting from a simple
  baseline, each \sys training component is added incrementally. Val MSE for each modality is averaged over the last 10 training epochs.}
  \label{table:ablation}
  \centering
  \begin{tabular}{c|l|ccc}
    \toprule
    Row & Configuration & EEG  & ECG  & EMG \\
    \midrule
    1. & 1-Branch, 1CB, MSE Phase + MSE $A$  & 1.509 & 1.956 & 1.996 \\
    2. & 4-Branch, 8CB, MSE Phase + MSE $A$ & 0.946 & 1.417 & 1.651\\
    \midrule
    3. & 4-Branch, 8CB, MSE Phase + $\log(A)$ + Temporal  & 0.603 & 0.843 & 1.000 \\
    4. & 4-Branch, 8CB, Phase Loss (without $\mathcal{L}_{\text{reg}}$) + MSE $A$ & 0.167 & 0.287 & 0.903 \\
    \midrule
     5. & 4-Branch, 8CB, Phase Loss (without $\mathcal{L}_{\text{reg}}$) + $\log(A)$  + Temporal & 0.035 & 0.115 & 0.447 \\
    \bottomrule
  \end{tabular}
\end{table}

Table~\ref{table:ablation} reports the validation MSE for each configuration 
across all three modalities. Each component contributes to reconstruction 
quality, with consistent trends across EEG, ECG, and EMG:

\begin{myenumerate}
    \item \textbf{Multi-scale architecture} (rows 1$\to$2): Scaling from 
    1 branch with 1 codebook to 4 branches with 8 codebooks per branch 
    reduces MSE by 37\% for EEG (1.509$\to$0.946), 28\% for ECG 
    (1.956$\to$1.417) and 17\% for EMG (1.996$\to$1.651), confirming 
    that decomposing the signal into frequency-specific representations 
    with deeper residual quantization leads to more faithful discrete 
    approximations across modalities.
    
    \item \textbf{Log-amplitude and temporal losses} (rows 2$\to$3): 
    Replacing raw amplitude MSE with the log-amplitude formulation 
    $\log(1+A)$ and adding the temporal reconstruction term reduces 
    EEG MSE from 0.946 to 0.603, ECG from 1.417 to 0.843 and EMG 
    from 1.651 to 1.000. The log transform compresses the dynamic range 
    and places greater emphasis on high-frequency components, while the 
    temporal loss provides complementary time-domain guidance.
    
    \item \textbf{Phase loss alone} (rows 2$\to$4): Replacing 
    the independent MSE phase loss with the cosine-similarity-based 
    $\mathcal{L}_{\text{phase}}$ (see Appendix~\ref{apx:sine/cosine}), 
    without any temporal or log-amplitude changes, substantially reduces 
    MSE for EEG (0.946$\to$0.167), ECG (1.417$\to$0.287) and EMG 
    (1.651$\to$0.903), demonstrating that respecting the circular 
    topology of phase is critical for reconstruction quality.
    
    \item \textbf{Full combination. Phase-Aware Loss} (rows 3$\to$5): Combining all 
    components produces the largest single-step improvement across all 
    modalities: EEG drops from 0.603 to 0.035 (17$\times$), ECG from 
    0.843 to 0.115 (7.3$\times$) and EMG from 1.000 to 0.447 (2.2$\times$). This dramatic gain demonstrates a synergy between 
    the phase and the temporal terms: the temporal term 
    provides stable time-domain gradients that enable the phase term to fully exploit its directional optimization.
\end{myenumerate}

In total, the full \sys configuration achieves a \textbf{43$\times$} reduction  in EEG reconstruction error (1.509$\to$0.035), \textbf{17$\times$} for ECG (1.956$\to$0.115) and \textbf{4.5$\times$} for EMG (1.996$\to$0.447) relative 
to the baseline. These results demonstrate that architectural contributions and training loss design are both essential and synergistic: while the multi-scale structure provides consistent gains, the largest improvements emerge from the phase-aware loss, underscoring that principled loss design is as important as model architecture for high-fidelity biosignal tokenization.

\subsection{Sensitivity Analysis}

The phase-aware loss (see Section \ref{sec:tokenizer_training_analysis}) follows a principled design grounded in signal processing. As shown in Appendix~\ref{apx:sine/cosine}, raw phase MSE is fundamentally flawed. We address this by representing phase as 2D vectors $(\cos\phi, \sin\phi)$ and use the cosine similarity to jointly optimise the predicted values $\hat{c}$ and $\hat{s}$. Although cosine  similarity is scale-invariant and does not explicitly constrain $(\hat{c}, \hat{s})$ to lie on the unit circle, the predicted norms tend to remain close to unity in practice (see Appendix \ref{apx:sine/cosine}). Nonetheless, as discussed, an optional regularization term can be added to explicitly penalize deviations from unit circle (see Appendix \ref{apx:sine/cosine}). To determine the sensitivity of reconstruction quality to $\lambda_{\text{circle}}$, we conduct an ablation across all three modalities with the full \sys architecture (4 branches, 8 codebooks per branch), varying $\lambda_{\text{circle}}$ from 0 (cosine similarity only, no unit-circle penalty) to 1.0 and using the datasets described in Section \ref{sec:scaling_analysis}. Table~\ref{table:lambda_ablation} reports the validation MSE for each setting.

\begin{table}[!h]
  \caption{Sensitivity analysis of the unit-circle penalty weight 
  $\lambda_{\text{circle}}$. Val MSE is averaged 
  over the last 10 training epochs.}
  \label{table:lambda_ablation}
  \centering
  \begin{tabular}{l|ccc}
    \toprule
    $\lambda_{\text{circle}}$ & EEG & ECG & EMG \\
    \midrule
    0.0 & 0.035 & 0.115 & 0.447 \\
    0.1 & 0.050 & 0.144 & 0.530 \\
    0.4 & 0.106 & 0.144 & 0.534 \\
    0.8 & 0.107 & 0.180 & 0.528\\
    1.0 & 0.111 & 0.143 & 0.525 \\
    \bottomrule
  \end{tabular}
\end{table}

Even without the unit-circle penalty ($\lambda_{\text{circle}}=0$), the cosine similarity alone achieves the lowest validation MSE across all modalities, already far below the MSE phase baselines (Table~\ref{table:ablation}), confirming that the core improvement comes from replacing independent phase MSE with directional alignment via cosine similarity. For EEG, increasing $\lambda_{\text{circle}}$ degrades reconstruction quality monotonically. For ECG and EMG, performance is relatively stable across $\lambda_{\text{circle}} \in [0.1, 1.0]$ but slightly worse than $\lambda_{\text{circle}} = 0.0$, indicating that stronger penalty values over-constrain the optimization, as discussed in Appendix \ref{apx:sine/cosine}. In our main results, however, we retain $\lambda_{\text{circle}} = 0.1$ as a conservative choice that provides an explicit geometric guarantee at small cost.

\section{Temporal Encoder Configuration \& Tokenizer Hyperparameter Settings}
\label{apx:hypers_tokenizer}

This section details: (\emph{i}) the configuration of \sys{}'s multi-scale temporal encoder (\S\ref{apx:arch_tokenizer}) for each modality, and (\emph{ii}) the hyperparameter settings used for training the tokenizer (\S\ref{apx:params_tokenizer}).

\subsection{Multi-Scale Temporal Encoder Architecture}
\label{apx:arch_tokenizer}

Our temporal encoder comprises four temporal branches. Each branch applies the following sequence of operations twice in succession (Figure~\ref{fig:neurorvq_tokenizer}): a 1-D convolution, followed by group normalization with $G=4$ groups and $C=8$ channels, GELU activation, and pooling. As described in Section \ref{sec:modality_config}, the signals for each modality are resampled to different sampling frequencies to maintain the necessary signal properties. Therefore, 
the temporal convolution kernels  are chosen to span receptive fields aligned  with the characteristic of each modality, with larger kernels broader morphological features and narrower kernels sensitive to capture the fast transients. Tables ~\ref{tab:temporal-config_eeg} and ~\ref{tab:temporal-config_emg} list the filter counts, kernel sizes, padding and pooling parameters for the two sequences, where $x \rightarrow y$ denotes the configuration value(s) used in the first and second sequence, respectively. Since both EEG and ECG modalities are resampled at the same rate (see Section \ref{sec:modality_config}), same kernel sizes are used.

\begin{table}[!h]
  \caption{Configuration of \sys's multi-scale temporal encoder for EEG and ECG modalities. Here, $x \rightarrow y$ indicates values for the first ($x$) and second ($y$) sequence.}
  \centering
    \begin{tabulary}{\linewidth}{ccccc}
    \toprule
    Branch \textnumero & Filters & Kernel & Padding & Pooling \\
    \midrule
    1 & 
    8 $\rightarrow$ 8 & 
    (1, 21) $\rightarrow$ (1, 9) & 
    (0, 10) $\rightarrow$ (0, 4) & 
    (1, 2) $\rightarrow$ (1, 4) \\
    2 & 
    8 $\rightarrow$ 8 & 
    (1, 15) $\rightarrow$ (1, 7) & 
    (0,~~~7) $\rightarrow$ (0, 3) & 
    (1, 2) $\rightarrow$ (1, 4) \\
    3 & 
    8 $\rightarrow$ 8 & 
    (1,~~~9) $\rightarrow$ (1, 5) & 
    (0,~~~4) $\rightarrow$ (0, 2) & 
    (1, 2) $\rightarrow$ (1, 4) \\
    4 & 
    8 $\rightarrow$ 8 & 
    (1,~~~5) $\rightarrow$ (1, 3) & 
    (0,~~~2) $\rightarrow$ (0, 1) & 
    (1, 2) $\rightarrow$ (1, 4) \\
    \bottomrule
  \end{tabulary}
  \label{tab:temporal-config_eeg}
\end{table}

\begin{table}[!h]
  \caption{Configuration of \sys's multi-scale temporal encoder for EMG modality. Here, $x \rightarrow y$ indicates values for the first ($x$) and second ($y$) sequence.}
  \centering
    \begin{tabulary}{\linewidth}{ccccc}
    \toprule
    Branch \textnumero & Filters & Kernel & Padding & Pooling \\
    \midrule
    1 & 
    8 $\rightarrow$ 8 & 
    (1, 51) $\rightarrow$ (1, 25) & 
    (0, 25) $\rightarrow$ (0, 12) & 
    (1, 2) $\rightarrow$ (1, 4) \\
    2 & 
    8 $\rightarrow$ 8 & 
    (1, 17) $\rightarrow$ (1, 9) & 
    (0,~~~8) $\rightarrow$ (0, 4) & 
    (1, 2) $\rightarrow$ (1, 4) \\
    3 & 
    8 $\rightarrow$ 8 & 
    (1,~~~8) $\rightarrow$ (1, 4) & 
    (0,~~~4) $\rightarrow$ (0, 2) & 
    (1, 2) $\rightarrow$ (1, 4) \\
    4 & 
    8 $\rightarrow$ 8 & 
    (1,~~~5) $\rightarrow$ (1, 3) & 
    (0,~~~2) $\rightarrow$ (0, 1) & 
    (1, 2) $\rightarrow$ (1, 4) \\
    \bottomrule
  \end{tabulary}
  \label{tab:temporal-config_emg}
\end{table}

\subsection{Tokenizer Hyperparameter Settings}
\label{apx:params_tokenizer}

Table \ref{tab:hyperparameters_appendix_tokenizer} summarises the hyperparameters used for training the tokenizer as described in Section \ref{sec:tokenizer_training_analysis}. Certain choices (e.g., batch size) were constrained by hardware, while others were guided by ablation studies and scaling analyses (see Section \ref{sec:scaling_analysis}).

\begin{table}[!h]
\caption{\sys Tokenizer Hyperparameters}
\centering
  \begin{tabular}{lccc}
    \toprule
    Hyperparameter & EEG & ECG & EMG\\
    \midrule
    Number of tokens in $\mathcal{V}_{i}$ & 8192 & 8192 & 8192\\
    Token Dimension & 128 & 128 & 128 \\
    Unit-Circle Penalty $\lambda_{circle}$ & 0.1 & 0.1 & 0.1
    \\
    Temporal Branches $S$ & 4 & 4 & 4 \\
    Number of codebooks $N$ per branch & 8 & 8 & 16 \\
    \midrule
    Batch size & 256 & 256 & 256\\
    Learning rate scheduler & Cosine & Cosine & Cosine \\
    Base learning rate & 5e-5 & 5e-5 & 5e-5\\
    Min learning rate & 1e-5  & 1e-5 & 1e-5\\
    Total epochs & 100 & 100 & 100\\
    Warmup epochs & 10 & 10  & 10 \\
    Optimizer & AdamW  & AdamW & AdamW\\
    Weight decay & 1e-4 & 1e-4 & 1e-4 \\
    Adam $\beta$ & (0.9, 0.999) & (0.9, 0.999) & (0.9, 0.999)\\   
    Gradient clipping & 3 & 3 & 3\\
    Layer scale init & 0.001 & 0.001 & 0.001\\
    Encoder depth & 12 & 12 & 12\\
    Decoder depth & 3  & 3  & 3 \\
    Hidden dimension & 200 & 40 & 200 \\
    No. Attention heads & 10 & 10 & 10\\
    \bottomrule
  \end{tabular}
  \label{tab:hyperparameters_appendix_tokenizer}
\end{table}

\newpage
\section{Scaling Analysis}
\label{apx:scaling_details}

\begin{figure}[!h]
    \centering
\includegraphics[width=\linewidth]{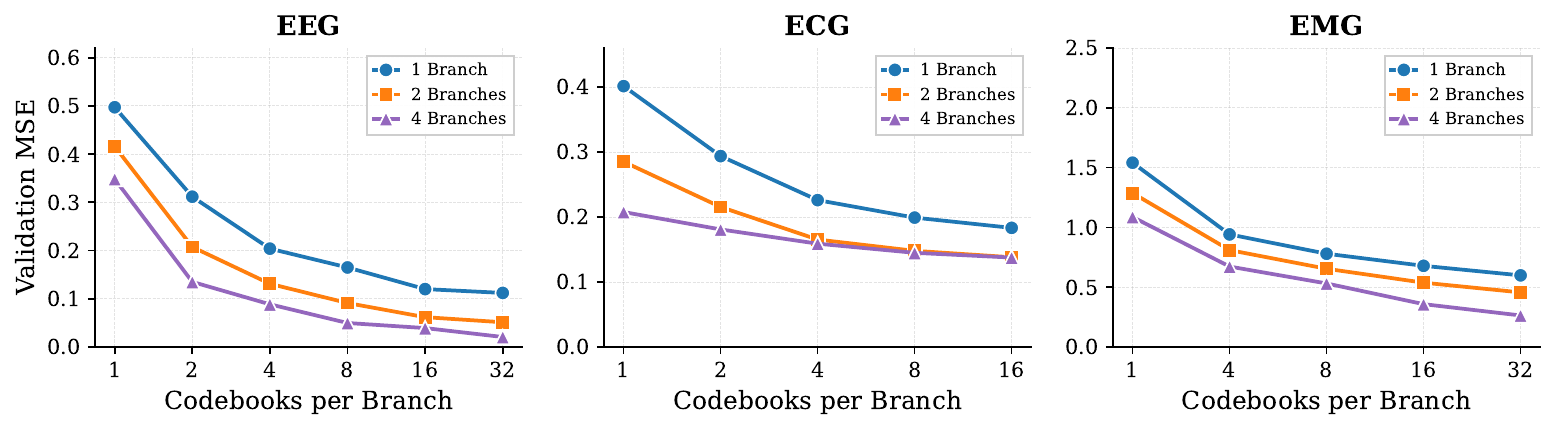}
    \caption{Mean squared error (MSE) reconstruction performance for \sys with various numbers of temporal branches and RVQ codebooks (CB) per branch across three modalities (EEG, ECG, EMG).}
\label{fig:scaling_mse}
\end{figure}

\begin{figure}[!h]
\centering
\includegraphics[width=\linewidth]{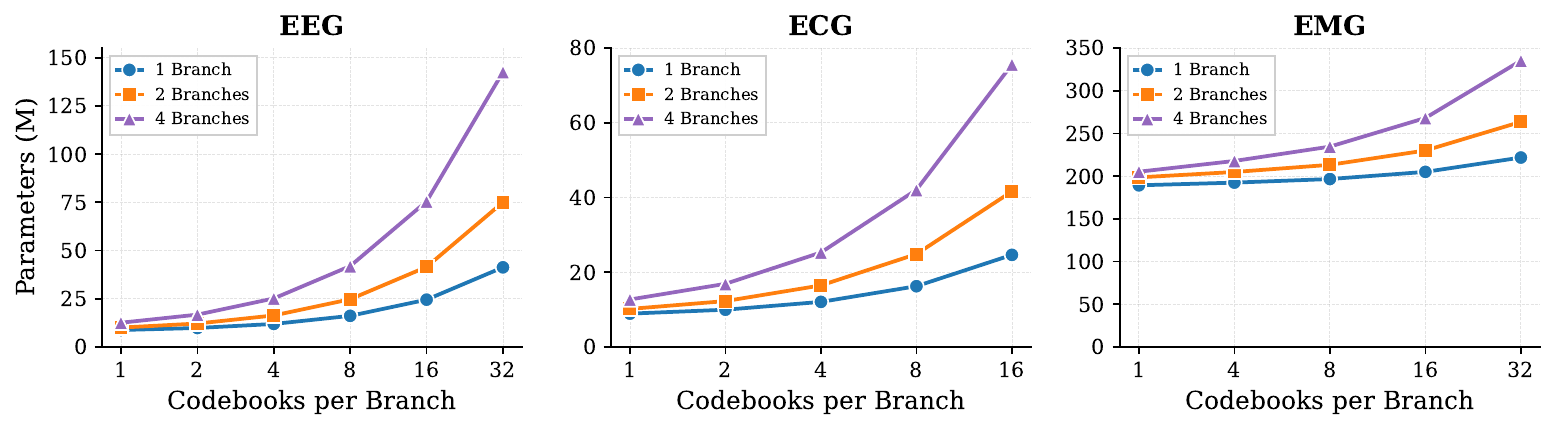}
    \caption{Number of trainable parameters for \sys with various numbers of temporal branches and RVQ codebooks (CB) per branch across three modalities (EEG, ECG, EMG).}
\label{fig:scaling_params}
\end{figure}

Figures ~\ref{fig:scaling_mse} and ~\ref{fig:scaling_params} provide the individual scaling analysis plots that are combined in Figure~\ref{fig:scaling_laws} in Section \ref{sec:modality_config}. 

Figure~\ref{fig:scaling_mse} shows the validation MSE as a function of codebook depth for each branch configuration, and Figure~\ref{fig:scaling_params} shows the corresponding number of trainable parameters. Together, these plots illustrate that reconstruction quality improves with both codebook depth and number of branches, while the parameter cost scales linearly with codebook depth and proportionally with the number of branches.

\newpage
\section{Codebook Utilization Analysis}
\label{apx:codebook_utilization}

Codebook utilization is an important property for an RVQ-based tokenizer. Using the held-out validation sets (Section \ref{sec:scaling_analysis}) per modality (EEG, ECG, EMG), the following metrics across all branches (4 for all modalities) and RVQ levels (8 for EEG/ECG and 16 for EMG) were computed: \emph{(i)} utilization, \emph{(ii)} perplexity, \emph{(iii)} dead codes, \emph{(iv)} top-1\% mass (selection probability captured by the 1\% most-used codes). The results are presented in Tables \ref{tab:eeg_rvq}, \ref{tab:emg_rvq} and \ref{tab:ecg_rvq}. \sys-tokenizer shows efficient code usage with no collapse: EEG and EMG reach 96.9\% and 99.2\% mean utilization with near-uniform usage and only 257 and 62 dead codes per 8192-entry codebook respectively. Within each modality the coarse first RVQ level has the lowest utilization while deeper residual levels saturate toward 100\%, with perplexity increasing and top-code concentration decreasing monotonically with depth. This behavior is a residual hierarchy: first levels produce coarse details and successive levels encode progressively finer detail spread across the full codebook. ECG shows lower utilization: this is expected and consistent with our scaling analysis (in Section \ref{sec:scaling_analysis}), which shows that ECG's stereotyped spectrally simple morphology saturates with far fewer codebooks than EEG or EMG.

\begin{table}[!h]
\centering
\begin{minipage}[t]{0.48\linewidth}
\centering
\caption{Codebook utilization for EEG}
\label{tab:eeg_rvq}
\resizebox{\linewidth}{!}{%
\begin{tabular}{lrrrr}
\toprule
RVQ level & Utilization & Perplexity & Dead codes & Top-1\% mass \\
\midrule
1 & 80.9\%  & 6099 & 1566 & 0.039 \\
2 & 97.1\%  & 7562 & 234  & 0.029 \\
3 & 97.7\%  & 7668 & 188  & 0.026 \\
4 & 99.5\%  & 7815 & 40   & 0.025 \\
5 & 100.0\% & 7846 & 3    & 0.024 \\
6 & 100.0\% & 7802 & 1    & 0.022 \\
7 & 99.9\%  & 7748 & 6    & 0.021 \\
8 & 99.8\%  & 7740 & 18   & 0.018 \\
\midrule
\textbf{Mean} & \textbf{96.9\%} & \textbf{7535} & \textbf{257} & \textbf{0.026} \\
\bottomrule
\end{tabular}}
\end{minipage}
\hfill
\begin{minipage}[t]{0.48\linewidth}
\centering
\caption{Codebook utilization for ECG}
\label{tab:ecg_rvq}
\resizebox{\linewidth}{!}{%
\begin{tabular}{lrrrr}
\toprule
RVQ level & Utilization & Perplexity & Dead codes & Top-1\% mass \\
\midrule
1 & 29.2\% & 2060 & 5798 & 0.123 \\
2 & 67.9\% & 4708 & 2631 & 0.068 \\
3 & 54.3\% & 3846 & 3746 & 0.101 \\
4 & 47.8\% & 3408 & 4280 & 0.100 \\
5 & 53.0\% & 3805 & 3848 & 0.085 \\
6 & 47.0\% & 3465 & 4342 & 0.091 \\
7 & 47.0\% & 3544 & 4338 & 0.079 \\
8 & 50.2\% & 3850 & 4076 & 0.060 \\
\midrule
\textbf{Mean} & \textbf{49.6\%} & \textbf{3586} & \textbf{4132} & \textbf{0.089} \\
\bottomrule
\end{tabular}}
\end{minipage}

\vspace{1em}

\begin{minipage}[t]{0.6\linewidth}
\centering
\caption{Codebook utilization for EMG}
\label{tab:emg_rvq}
\resizebox{\linewidth}{!}{%
\begin{tabular}{lrrrr}
\toprule
RVQ level & Utilization & Perplexity & Dead codes & Top-1\% mass \\
\midrule
1  & 88.1\%  & 6442 & 978 & 0.053 \\
2  & 100.0\% & 7331 & 1   & 0.049 \\
3  & 99.9\%  & 7355 & 5   & 0.048 \\
4  & 100.0\% & 7388 & 0   & 0.046 \\
5  & 100.0\% & 7418 & 2   & 0.044 \\
6  & 100.0\% & 7454 & 0   & 0.042 \\
7  & 100.0\% & 7488 & 1   & 0.039 \\
8  & 100.0\% & 7517 & 0   & 0.037 \\
9  & 100.0\% & 7541 & 0   & 0.036 \\
10 & 100.0\% & 7568 & 0   & 0.035 \\
11 & 100.0\% & 7618 & 0   & 0.034 \\
12 & 100.0\% & 7632 & 0   & 0.034 \\
13 & 100.0\% & 7733 & 0   & 0.032 \\
14 & 100.0\% & 7812 & 0   & 0.030 \\
15 & 100.0\% & 7844 & 0   & 0.028 \\
16 & 100.0\% & 7913 & 0   & 0.027 \\
\midrule
\textbf{Mean} & \textbf{99.2\%} & \textbf{7503} & \textbf{62} & \textbf{0.038} \\
\bottomrule
\end{tabular}}
\end{minipage}
\end{table}

\newpage
\section{Datasets}
\label{apx:dataset}

We use a collection of publicly available datasets to pre-train \sys for each biosignal modality. Tables ~\ref{dataset-table_emg}, ~\ref{dataset-table_ecg}, ~\ref{dataset-table_eeg} summarize the datasets, their associated BCI paradigms and the number of sensors. The collection spans multiple experimental settings, including hand pose estimation, touch typing, cardiac arrhythmia detection, motor imagery, event-related potentials (ERP), resting-state activity, information-integration and rule-based cognitive categorization (II-RB) and clinical data for epilepsy and seizures.

\begin{table}[!h]
  \caption{Datasets used for pre-training \sys in EMG.}
  \label{dataset-table_emg}
  \centering
  \begin{tabular}{lcc}
    \toprule
    Datasets & Task  & \# channels \\
    \midrule
    emg2pose~\citep{salteremg2pose} & Hand Pose Estimation & 16 \\
    emg2qwerty~\citep{emg2qwerty} & Touch Typing & 16 \\
    \bottomrule
  \end{tabular}
\end{table}

\begin{table}[!h]
  \caption{Datasets used for pre-training \sys in ECG.}
  \label{dataset-table_ecg}
  \centering
  \begin{tabular}{lcc}
    \toprule
    Datasets & Task  & \# leads \\
    \midrule
    PTB~\citep{Bousseljot2004-wi} & Diagnostic Classification & 15 \\
    MIMIC-IV-ECG~\citep{mimicivecg} & Diagnostic Classification & 12 \\
    Chapman-Shaoxing~\citep{chapman} & Arrhythmia Classification & 12 \\
    \bottomrule
  \end{tabular}
\end{table}

 \begin{table}[!h]
  \caption{Datasets used for pre-training \sys in EEG.}
  \label{dataset-table_eeg}
  \centering
  \begin{tabular}{lcc}
    \toprule
    Datasets & BCI paradigm  & \# channels \\
    \midrule
    BCI Competition IV-1~\citep{bci_iv1} &  Motor & 64 \\
    Grasp and Lift~\citep{grasplift}  & Motor & 32 \\
    Physionet MI~\citep{physionetmi} & Motor & 64\\
    bi2015a~\citep{bi2015a} & ERP & 32 \\
    Inria BCI Challenge~\citep{inriaBCI} & ERP & 56 \\
    \citet{Trujillo2020} & II-RB & 64 \\
    \citet{Trujillo2017} & Resting & 64\\
    SPIS Resting~\citep{spisresting} & Resting & 64 \\
    TUAR~\citep{TUAR} & Artifacts (e.g., eye movement) & 31 \\
    Siena Scalp~\citep{sienascalp} & Epilepsy & 31 \\
    TUEP~\citep{TUEP} & Epilepsy & 31\\
    TUSZ~\citep{TUSZ} & Seizure & 31\\
    \midrule
    Self-Collected Dataset & Motor & 29 \\
    \bottomrule
  \end{tabular}
\end{table}

In the EEG modality, we additionally include a proprietary dataset comprising approximately 235 hours of motor imagery data recorded with 29 channels. The data collection was approved by the Research Governance and Integrity Team of Imperial College London (approval number 22IC7505). Recordings used non-invasive scalp EEG. All participants received the study information below and gave informed consent before taking part: 

\begin{quote}
``We will record your brain waves while you engage in simple tasks. These include real and imagined movements of the extremities, and a game which uses movements as controls. The collected data will be used to train algorithms that can be used for many things, such as facilitating navigation in virtual environments, seamless user interaction with objects in computer gaming or even rehabilitation training of post-stroke patients towards regaining the ability to control their limbs. The project received approval from the Research Governance and Integrity Team. If you want to help out our research team you can register to be a participant in our experiment! There is a small gift of £30 to every participant.''
\end{quote}

\newpage
\section{\sys Reconstruction Examples}
\label{apx:validation_curves}

In this section, we present a few reconstruction examples using the \sys tokenizer across all three modalities (EEG, ECG and EMG).

\begin{figure}[!h]
  \centering
\includegraphics[width=0.9\linewidth]{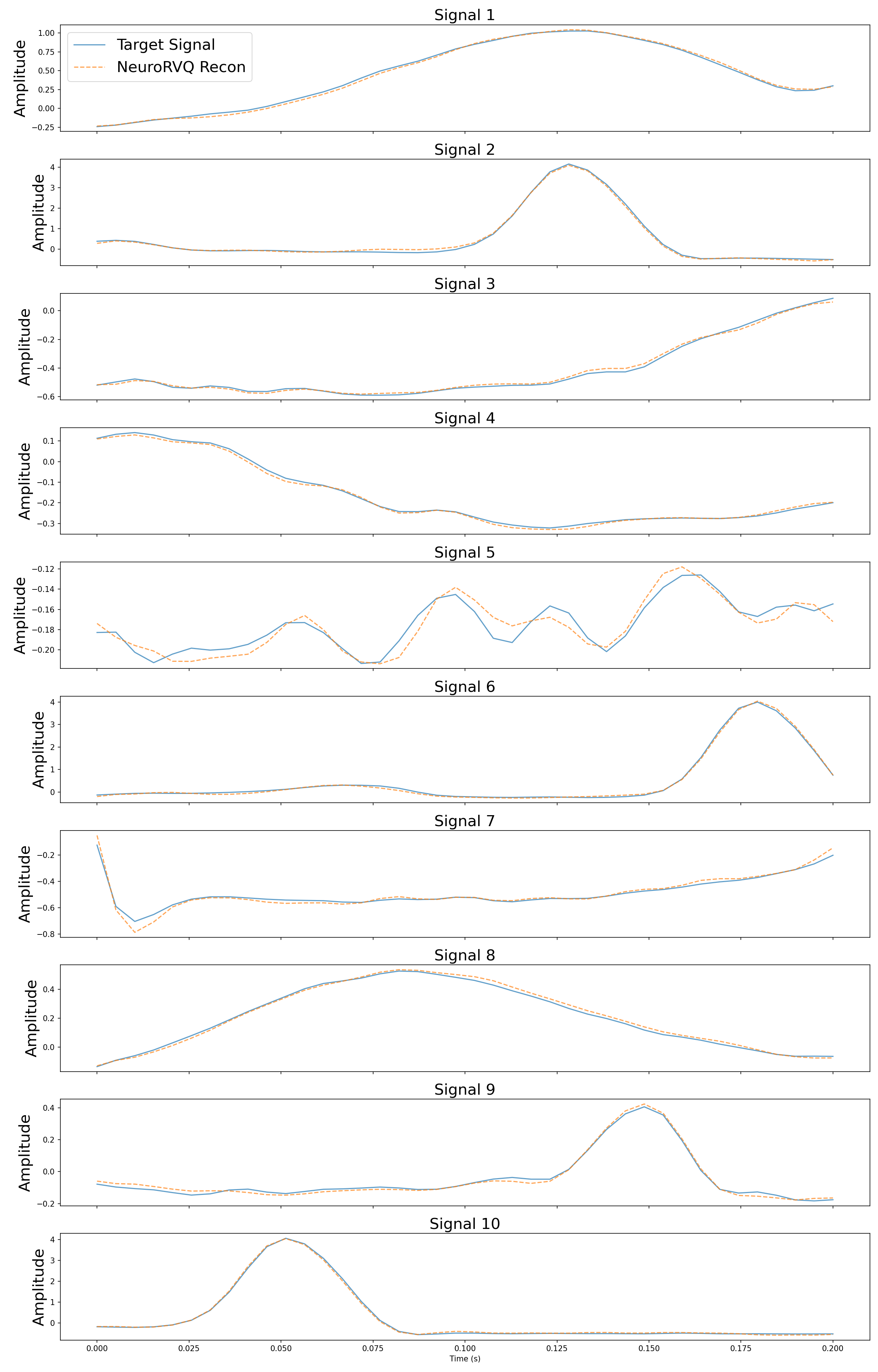}
  \caption{Reconstructed ECG signals (from the validation set) using \sys codebook-based tokenizer. \textcolor{blue}{Blue} lines denote the input ECG signal and \textcolor{orange}{orange} the reconstructed ECG signal.}
  \label{fig:reconstructions_large_ecg}
\end{figure}

\newpage
\begin{figure}[!h]
  \centering
\includegraphics[width=0.9\linewidth]{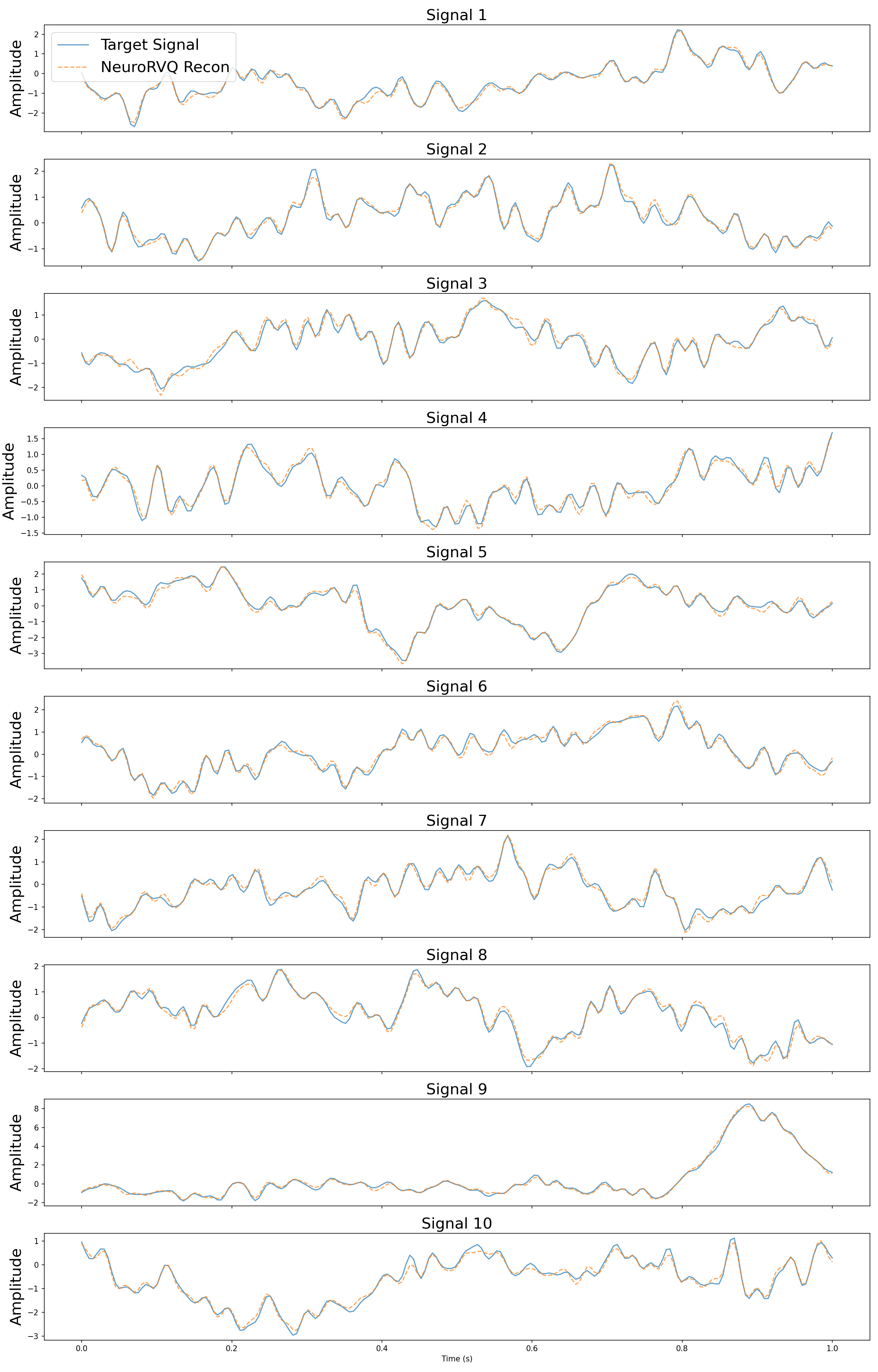}
  \caption{Reconstructed EEG signals (from the validation set) using \sys codebook-based tokenizer. \textcolor{blue}{Blue} lines denote the input EEG signal and \textcolor{orange}{orange} the reconstructed EEG signal.}
  \label{fig:reconstructions_large_ecg}
\end{figure}

\newpage
\begin{figure}[!h]
  \centering
\includegraphics[width=0.9\linewidth]{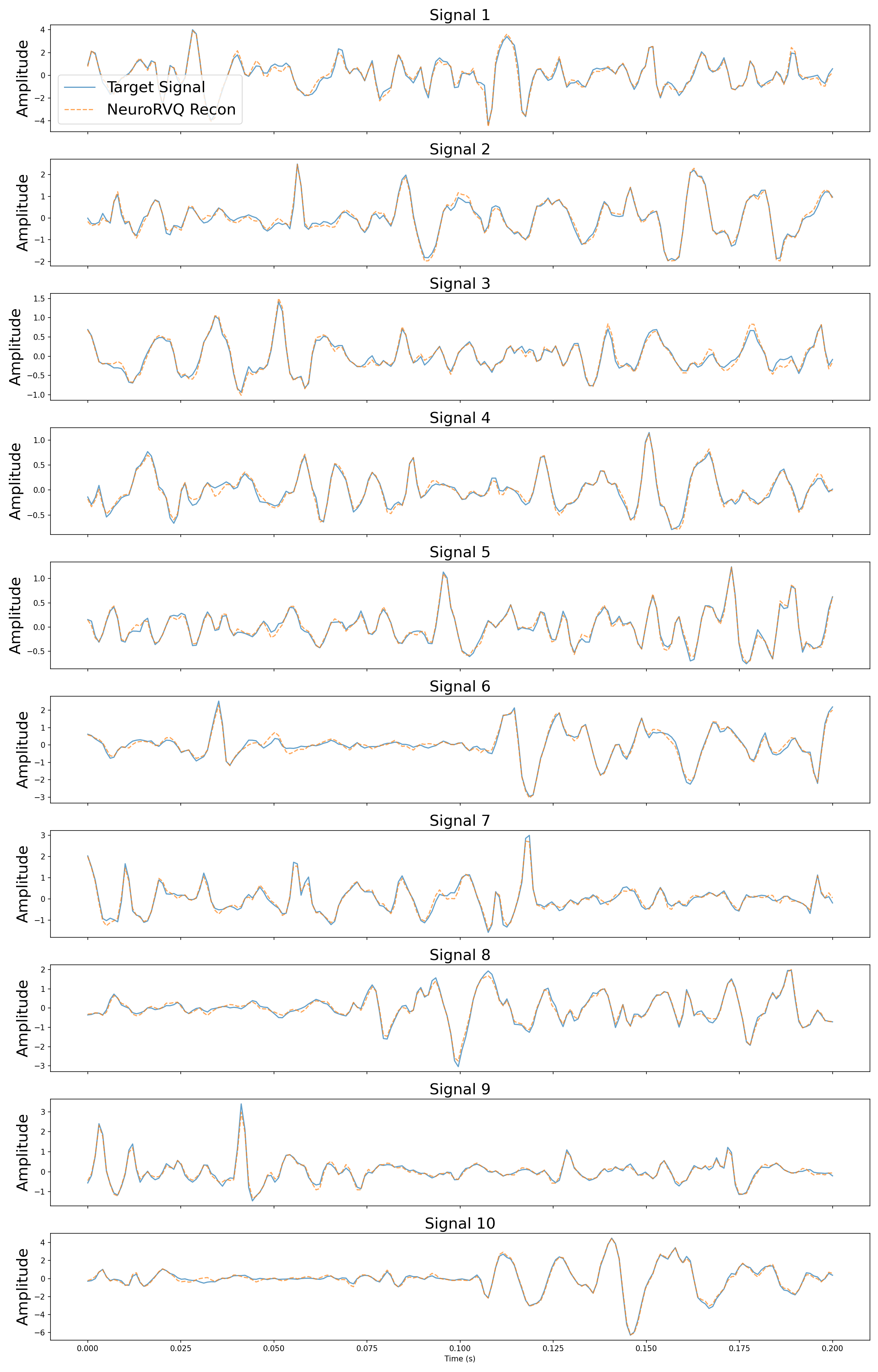}
  \caption{Reconstructed EMG signals (from the validation set) using \sys codebook-based tokenizer. \textcolor{blue}{Blue} lines denote the input EMG signal and \textcolor{orange}{orange} the reconstructed EMG signal.}
  \label{fig:reconstructions_large_ecg}
\end{figure}

\newpage
\begin{figure}[!h]
    \centering
    \begin{subfigure}[t]{0.48\linewidth}
        \centering
        \includegraphics[width=\linewidth]{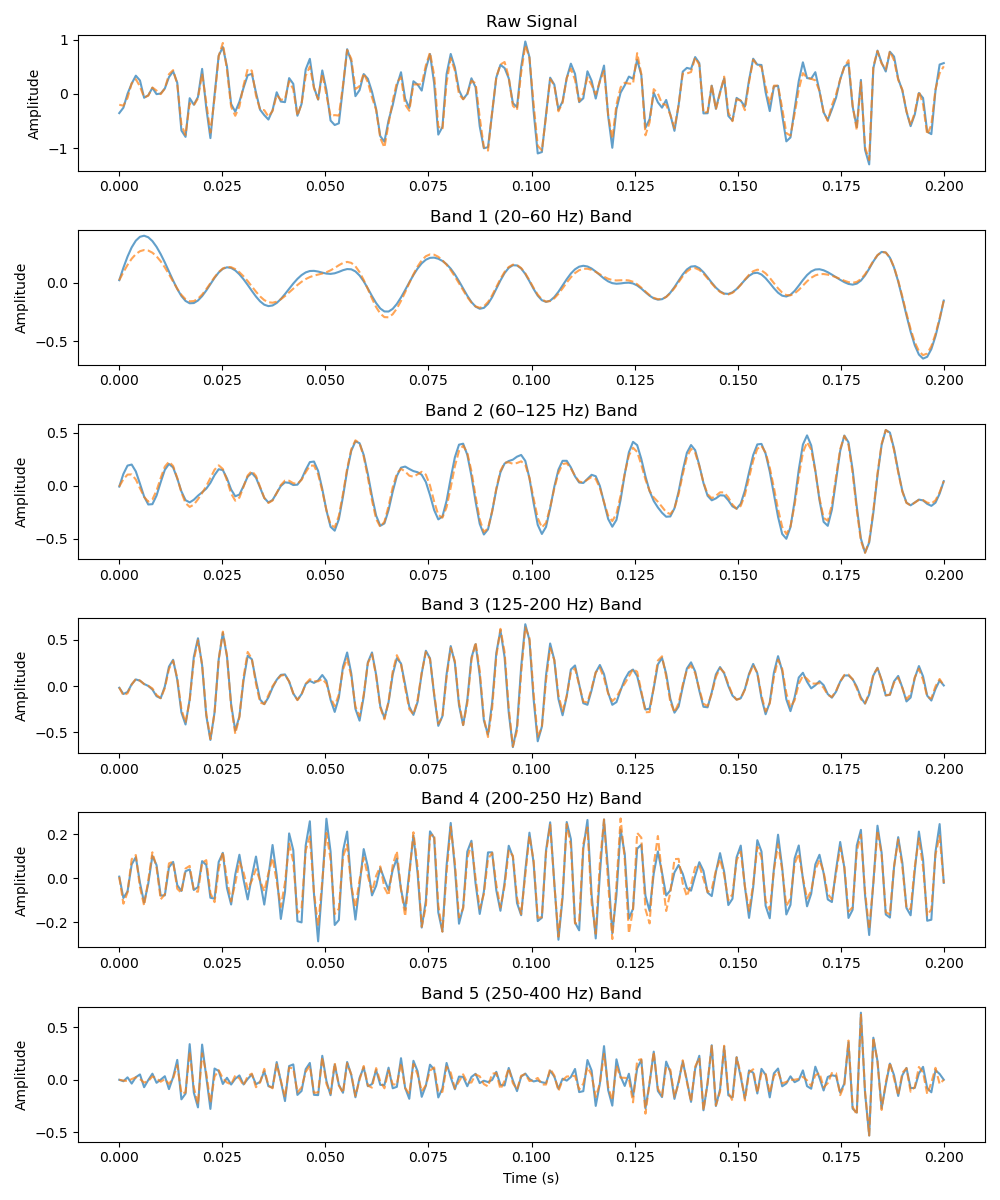}
    \end{subfigure}
    \hfill
    \begin{subfigure}[t]{0.48\linewidth}
        \centering
\includegraphics[width=\linewidth]{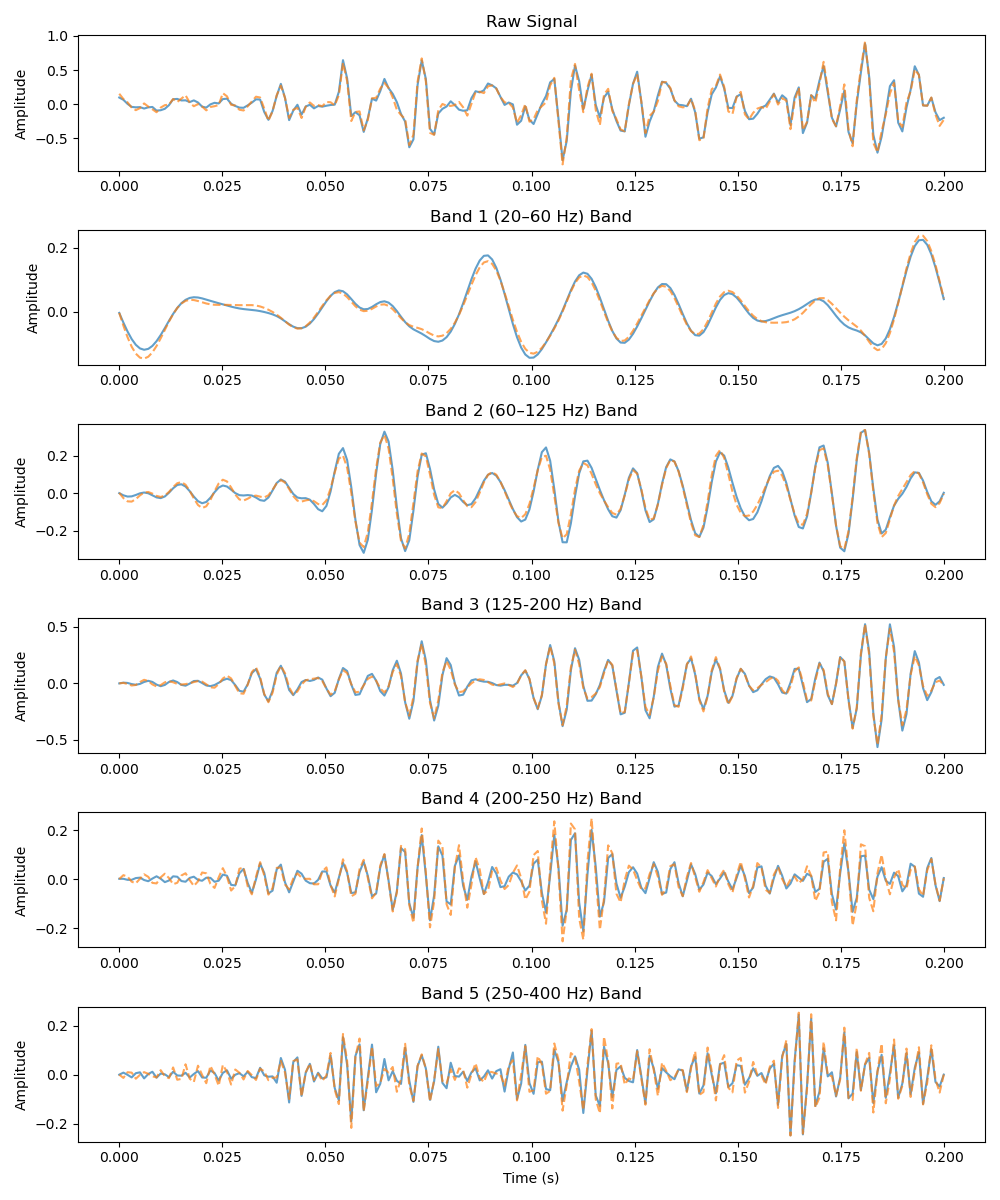}
    \end{subfigure}
\caption{Per-band analysis of two sample reconstructed EMG signal (from the validation set) from \sys codebook-based tokenizer. \textcolor{blue}{Blue} lines denote the input EMG signal and \textcolor{orange}{orange} the reconstructed EMG signal.}
\label{fig:reconstructions_emg_small}
\end{figure}

\begin{figure}[!h]
    \centering
    \begin{subfigure}[t]{0.48\linewidth}
        \centering
\includegraphics[width=\linewidth]{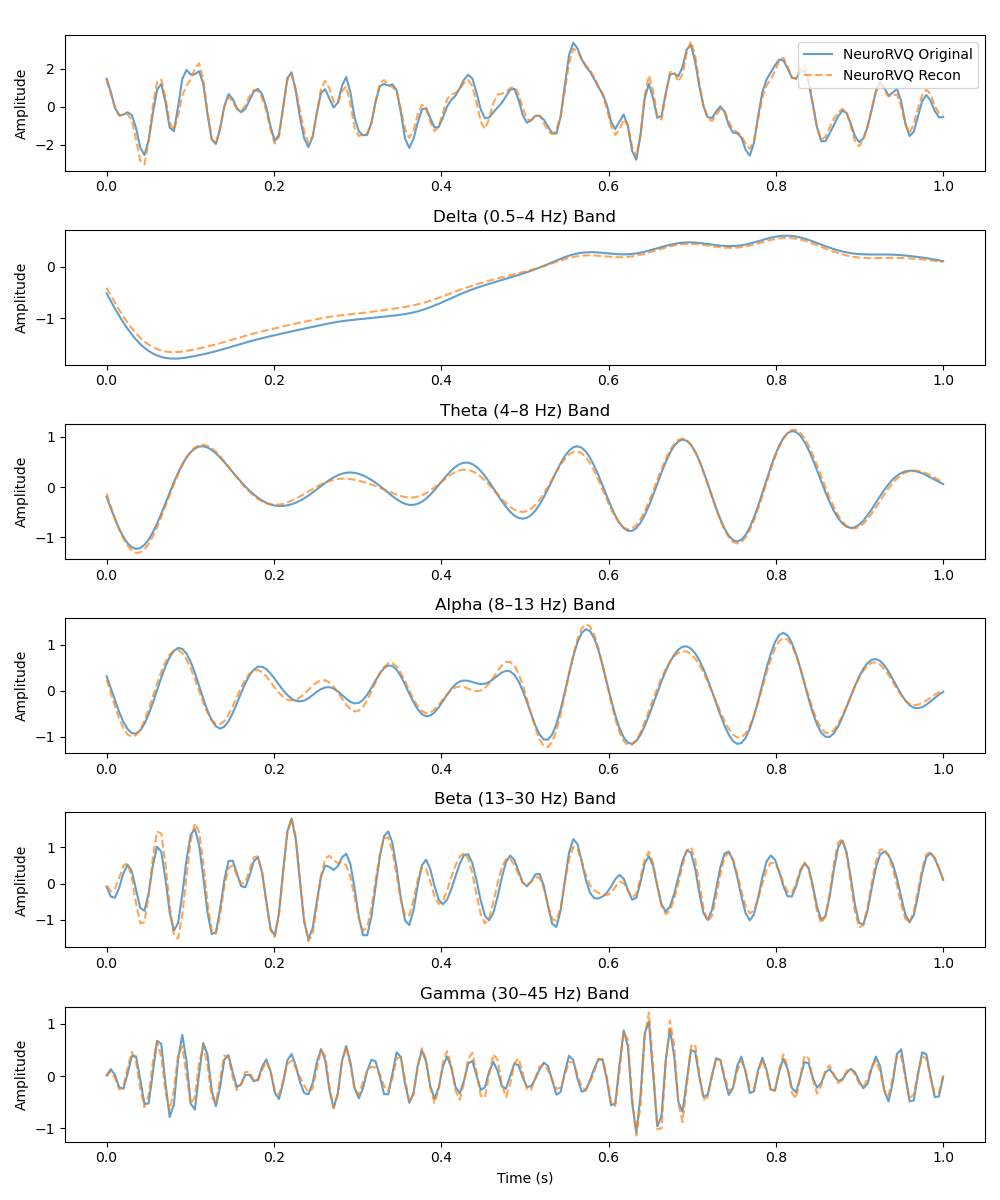}
    \end{subfigure}
    \hfill
    \begin{subfigure}[t]{0.48\linewidth}
        \centering
\includegraphics[width=\linewidth]{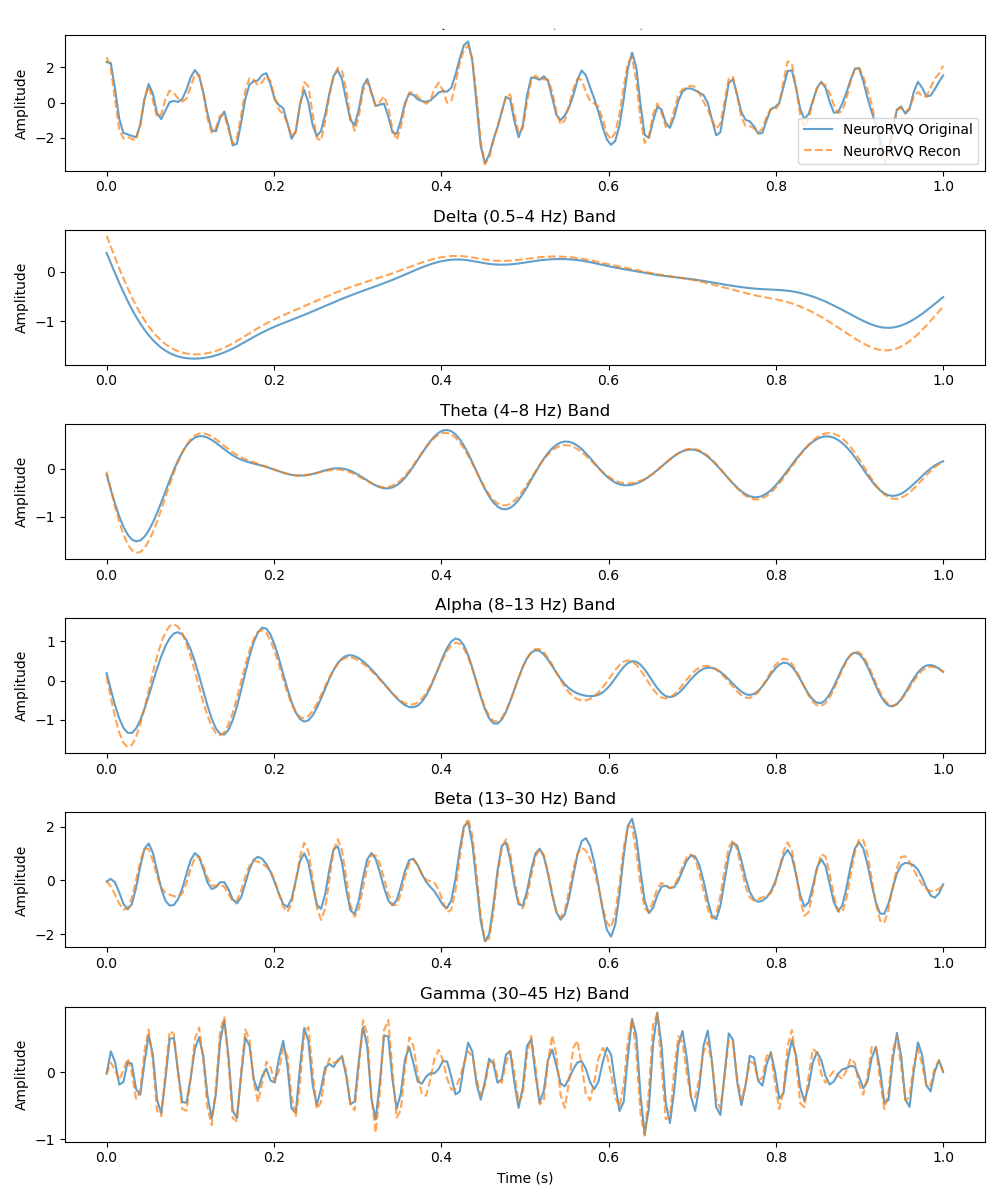}
    \end{subfigure}
\caption{Per-band analysis of two sample reconstructed EEG signal (from the validation set) from \sys codebook-based tokenizer. \textcolor{blue}{Blue} lines denote the input EEG signal and \textcolor{orange}{orange} the reconstructed EEG signal.}
\label{fig:reconstructions_eeg_small}
\end{figure}

\newpage

\section{\sys Comparison with Pre-Trained Codebook-Based Tokenizers}
\label{apx: other_tokenizers}

For EEG, we compare the signal reconstruction capabilities of \sys against two pretrained open-source state-of-the-art codebook-based tokenizers: LaBraM~\citep{LaBraM} and BrainOmni~\citep{xiao2025brainomni}. For ECG and EMG, no unimodal codebook-based tokenizers currently exist; \sys is the first to demonstrate high-fidelity codebook-based reconstruction for these modalities, as shown in the validation curves in Figure~\ref{fig:val_curves_all} and in Appendix~\ref{apx:validation_curves}. 

\paragraph{Comparison with LaBraM.} LaBraM employs a VQ-VAE-inspired tokenizer~\citep{esser2020taming} with a single codebook. We evaluate both tokenizers on an unseen dataset during training: the High Gamma motor dataset~\citep{HighGamma}. To compare both reconstructions, both signal are brought to the time domain and z-scored per trial together with the input before computing MSE, so the comparison is scale-matched. However, LaBraM is pre-trained between 0.1–75 Hz (LaBraM-Setup) while \sys is pre-trained between 0.5–45 Hz (\sys-Setup). We performed both setup comparisons in the motor-trials for fair comparison. As shown in Table~\ref{table:mse_results}, \sys consistently outperforms LaBraM across both setups, demonstrating superior reconstruction generalization.


\begin{table}[!h]
  \caption{Out-of-distribution MSE reconstruction performance for \sys and LaBraM tokenizers with both setups. \sys-Setup: 0.5-45Hz. LaBraM-Setup: 0.1-75Hz.}
  \centering
  \begin{tabulary}{\linewidth}{Lcc}
    \toprule
     & LaBraM-Setup & \sys-Setup \\
    \midrule
    LaBraM & 0.598 &  0.855\\
    \midrule
    \sys  & \textbf{0.072} & \textbf{0.058} \\
    \bottomrule
  \end{tabulary}
  \label{table:mse_results}
\end{table}



To verify that these gains stem from architectural differences rather than training data, we also conducted an in-distribution comparison in which LaBraM and \sys were trained on the same datasets, with identical train-validation splits and training regime (see Appendix~\ref{apx:in_distribution_curves}). The results confirm that \sys's advantage is consistent regardless of the training data used.

\begin{figure}[!h]
  \centering
  \includegraphics[width=\linewidth]{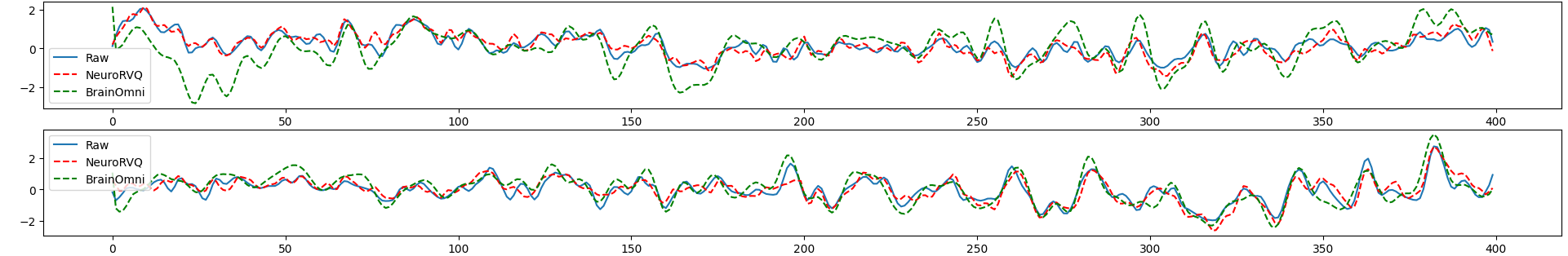}
  \caption{Two sample reconstructed EEG signals from BrainOmni and \sys codebook-based tokenizers. \textcolor{blue}{Blue} lines denote the input EEG signal, \textcolor{green}{green} the reconstructed EEG signal for BrainOmni and \textcolor{red}{red} the reconstructed EEG signal for \sys.}
  \label{fig:reconstructions_brainomni}
\end{figure}

\paragraph{Comparison with BrainOmni.} BrainOmni adopts a single residual quantizer inspired by EnCodec~\citep{encodec}, whereas \sys employs multiple RVQ codebooks, one per frequency scale. 
BrainOmni is pre-trained between 0.1–96 Hz
(BrainOmni-Setup) and uses different signal normalization strategy than \sys. Therefore, we evaluate on BrainOmni's publicly available sample data, which we use because their model requires device-specific layout information and we report results under both setups to ensure a fair comparison. As shown in Table~\ref{table:mse_brainomni} and Figure~\ref{fig:reconstructions_brainomni}, \sys consistently outperforms BrainOmni under both normalization conditions.

\begin{table}[!h]
  \caption{Out-of-distribution MSE reconstruction performance for \sys and BrainOmni tokenizers with both setups. \sys-Setup: 0.5-45Hz and \sys's normalization scheme. BrainOmni-Setup: 0.1-96Hz and BrainOmni's normalization scheme.}
  \centering
  \scalebox{0.90}{
  \begin{tabulary}{\linewidth}{Lcc}
    \toprule
     & \sys-Setup & BrainOmni-Setup \\
    \midrule
    BrainOmni & 0.350  & 0.413  \\
    \midrule
    \sys & \textbf{0.219} & \textbf{0.324}\\
    \bottomrule
  \end{tabulary}
  }
  \label{table:mse_brainomni}
\end{table}

\newpage
\section{\sys vs. LaBraM: Reconstruction Performance}
\label{apx:in_distribution_curves}

\begin{figure}[!h]
  \centering
\includegraphics[width=0.80\linewidth]{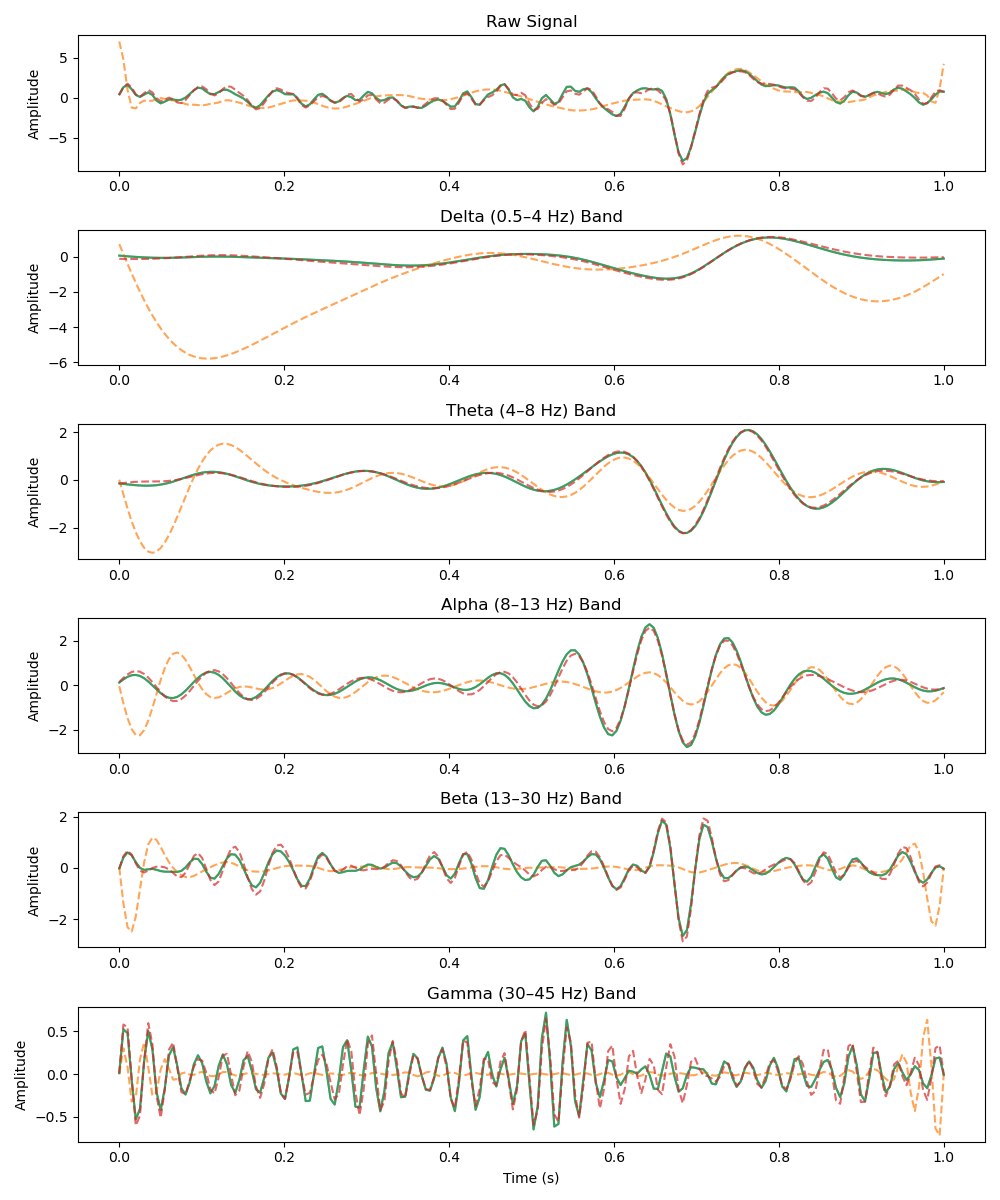}
  \caption{Per-band analysis of a sample reconstructed EEG signal (from the validation set) from LaBraM and \sys codebook-based tokenizers. \textcolor{green}{Green} lines denote the input EEG signal, \textcolor{orange}{orange} the reconstructed EEG signal for LaBraM (ours) and \textcolor{red}{red} the reconstructed EEG signal for \sys.}
  \label{fig:reconstructions}
\end{figure}

We train \sys and LaBraM tokenizers on \textbf{the same datasets with the same train–validation split} and an identical training regime (including preprocessing and training steps), as described in Section \ref{sec:setup_tokenizer}. As shown in Table \ref{table:mse_results_in_dist} and Figure \ref{fig:reconstructions}, \sys achieves orders of magnitude lower in-distribution reconstruction mean squared error (MSE) compared to LaBraM.

\begin{table}[!h]
  \caption{In-distribution validation mean squared error (MSE) reconstruction performance across frequency bands for \sys and LaBraM tokenizers trained on the same datasets.}
  \centering
  \begin{tabulary}{\linewidth}{L|cccccc}
    \toprule
    & Raw Signal & Delta & Theta & Alpha & Beta & Gamma \\
    \midrule
    LaBraM (ours) & 1.071 & 1.561 & 0.184 & 0.099 & 0.122 & 0.020 \\
    \sys  & \textbf{0.016} & \textbf{0.006} & \textbf{0.002} & \textbf{0.002} & \textbf{0.005} & \textbf{0.002} \\
    \bottomrule
  \end{tabulary}
  \label{table:mse_results_in_dist}
\end{table}

\newpage
\section{Foundation Model Configuration \& Hyperparameter Settings}
\label{apx:hypers_foundation}

This section details the hyperparameter settings used for pre-training the foundation model as well as its fine-tuning on 
downstream tasks. \sys foundation model used the multi-scale temporal encoder configurations described in (\S\ref{apx:arch_tokenizer}).

\begin{table}[!h]
\caption{\sys-FM Hyperparameters. The rightmost column reports the fine-tuning
protocol used for our 3-seed runs on Discrete Gestures and Wrist Velocity Regression.}
\centering
\small
  \begin{tabular}{
  >{\raggedright\arraybackslash}p{3.5cm}
  >{\centering\arraybackslash}p{2.5cm}
  >{\centering\arraybackslash}p{2cm}
  >{\centering\arraybackslash}p{2cm}
  >{\centering\arraybackslash}p{2.6cm}
  }
    \toprule
    Hyperparameter  & Pre-training FM (EEG/ECG/EMG) & Finetuning (EEG/ECG) & Finetuning (EMG) & Finetuning (EMG: \citep{Kaifosh2025-fi}) \\
    \midrule
    Number of tokens in $\mathcal{V}_{i}$ & 8192 & 8192 & 8192 & 8192 \\
    Token Dimension & 128 & 128 & 128 & 128 \\
    Unit-Circle Penalty $\lambda_{circle}$ & 0.1 & 0.1 & 0.1 & 0.1 \\
    Temporal Branches $S$ & 4 & 4 & 4 & 4 \\
    Number of codebooks $N$ & 8 & 8 & 16 & 16 \\
    \midrule
    Batch size & 64 & 64 & 128 & 64 \\
    Learning rate scheduler & Cosine & Linear & Cosine & Cosine \\
    Base learning rate & 5e-4 & 5e-4 & 1e-3 & 5e-4 \\
    Min learning rate & 1e-5 & - & 1e-6 & 0 \\
    Warmup lr start-end factors & - & (0.1, 1) & (0.01, 1.0) & (0.1, 1.0) \\
    Lr start-end factors & - & (1, 0.1) & (1, 0.001) & - \\
    Total epochs & 50 & 20 & 100 & 15 (Gestures) / 20 (Wrist) \\
    Warmup epochs & 5 & 4 & 5 & 3 \\
    Optimizer & AdamW & AdamW & AdamW & AdamW \\
    Weight decay & 0.05 & 0.01 & 0.001 & 0.01 \\
    Adam $\beta$ & (0.9, 0.999) & (0.9, 0.999) & (0.9, 0.95) & (0.9, 0.999) \\
    Layer lr decay & - & 0.975 & - & - \\
    Masking Ratio & 0.5 & - & - & - \\
    Layer scale init & 0.001 & - & - & - \\
    Encoder depth & 12 & 12 & 12 & 12 \\
    Hidden dimension & 200/40/200 & 200/40 & 200 & 200 \\
    No. Attention heads & 10 & 10 & 10 & 10 \\
    \bottomrule
  \end{tabular}
  \label{tab:hyperparameters_appendix}
\end{table}

Table \ref{tab:hyperparameters_appendix} summarises the hyperparameters used for pre-training and fine-tuning the foundation model. Certain choices (e.g., batch size) were constrained by hardware, while others were guided by ablation studies (see Section \ref{sec:scaling_analysis}).




\section{Downstream Task Fine-Tuning Analysis}
\label{apx:downstream_finetuning_analysis}

In this section, we describe the baseline foundation models, datasets / benchmarks and implementation details for each modality. Full hyperparameter configurations are provided in Appendix~\ref{apx:hypers_foundation}.

\subsection{EEG downstream evaluation.}
For EEG, we compare against nine foundation models on five downstream classification tasks from the benchmark of~\citep{lee2025assessing}. These models include two models that use codebook-based tokenizer like \sys, namely LaBraM ~\citep{LaBraM} and BrainOmni ~\citep{xiao2025brainomni} and seven common foundation model baselines, namely NeuroGPT\citep{NeuroGPT}, CBraMod~\citep{wang2025cbramod}, EEGPT~\citep{wang2024eegpt}, BIOT~\citep{NEURIPS2023_f6b30f3e},  MIRepNet~\citep{liu2025mirepnetpipelinefoundationmodel}, CSBrain~\citep{zhou2025csbrain}, REVE~\citep{ouahidi2025reve}. The five downstream classification tasks include: (\emph{a})~Motor, four-class movement classification on High Gamma~\citep{HighGamma}; (\emph{b})~ERP, binary event-related potential classification on Korea University~\citep{KU_ERP}; (\emph{c})~Memory, binary working memory classification on~\citet{Pavlov22}; (\emph{d})~Sleep, six-class sleep stage classification on Sleep-EDF~\citep{SleepEDF}; and (\emph{e})~Eyes, eyes-open vs.\ closed classification on PhysioNet Motor~\citep{physionetmi}. This benchmark was selected for its diverse BCI paradigms and minimal spurious artifacts. All models used their original pre-processing settings (for \sys-EEG 0.5-45Hz band pass filter), were fully fine-tuned for 20 epochs (early stopping of patience 5 epochs was employed to prevent overfitting), evaluated via 10-fold subject-independent cross-validation and compared using balanced accuracy.
As shown in Table~\ref{tab:downstream_results_all_modalities} (Section \ref{sec:downstream}), \sys-EEG achieves the best balanced accuracy on four of five tasks and the highest mean balanced accuracy overall. Notably, \sys uses only 5.9M backbone parameters, substantially fewer than other models, e.g. NeuroGPT (79.5M). This combination of strong generalization and compact size underscores that the tokenizer, not model scale, is the primary driver of performance. 

\paragraph{Ablation studies.} To ensure that our downstream gains are attributable to tokenization differences rather dataset selection or model architecture, we conduct two ablation studies: \emph{(i)} we follow the same controlled comparison protocol used for the tokenizer evaluation (Appendix~\ref{apx:in_distribution_curves}) where both \sys and LaBraM foundation models are trained on identical datasets with the same train--validation splits, preprocessing and training regime and \emph{(ii)} a tokenizer swap ablation. The results, reported in Appendix~\ref{apx:downstream_codebook} and Appendix ~\ref{apx:swap_ablation}, confirms that high fidelity tokenizer translates to better downstream task performance under matched conditions.


\subsection{ECG downstream evaluation.} For ECG, we compare against two foundation models, HuBERT-ECG~\citep{HuBERT} and ECGFounder~\citep{ECGFounder}, on the PTB-XL benchmark~\citep{PTB-XL-Nature}. PTB-XL is the most popular downstream dataset for ECG foundation models. It includes various labels covering Diagnostic, Form and Rhythm statements, but most models choose to use only a subset of these for multi-label classification. Following the structure of the original dataset \cite{PTB-XL-Nature}, we evaluated models on two established ECG classification tasks covering different levels of diagnostic granularity: (i) a 5-class task based on the coarse diagnostic superclasses and (ii) a 43-class task using the full set of fine-grained diagnostic statements.  We adopted the dataset’s recommended 10-fold, patient-stratified splitting scheme, using the designated nine folds for model training and validation and the reserved tenth fold for final testing. In this way, we followed a unified benchmarking protocol that allows fair performance comparisons across all evaluated ECG foundation models. All models were fully fine-tuned for 20 epochs (with early stopping to prevent overfitting). Table \ref{tab:downstream_results_all_modalities} reports metrics for each classification task for open-source ECG foundation models.
\sys achieves high balanced accuracy, doubling the BAcc of the next-best model on the fine-grained 43-class task, but ECGFounder leads on 5-class task. 

\subsection{EMG downstream evaluation.} 

We compared the performance of \sys-EMG against two foundation models, PhysioWave~\citep{chen2025physiowave} and TinyMyo~\citep{fasulo2026tinymyotinyfoundationmodel}, on four downstream hand gesture classification tasks: i) Discrete Gestures~\cite{Kaifosh2025-fi}: a 9-class gesture classification task using 2s EMG signals recorded using a 16 channel wristband at 2000 Hz ii) EPN-612~\citep{benalcazarEMGEPN612Dataset2020}: a 6-class classification task from 5s EMG signals recorded with an 8-channel EMG armband iii) NinaPro DB5~\citep{pizzolatoComparisonSixElectromyography2017}: A 52-class classification task from 5s EMG windows, recorded with two, 8-channel sEMG bands placed laterally adjacent on the forearm, iv) UCI-EMG~\citep{emg_data_for_gestures_481}: a 6-class classification from 1s windows of sEMG recorded using an 8-channel sEMG armband. 

Unlike previous sEMG foundation model evaluations, we exclusively split datasets at the subject level into train, validation and test sets with ratios 7:1:2 (for the \citep{Kaifosh2025-fi} dataset, we used a 7:1.5:1.5). This difference contributes to a lower performance across models, especially on the NinaPro DB5 dataset (Table~\ref{table:downstream_all}), compared to other works which train and test on different gesture repetitions within the same recording session. Another key difference on the NinaPro DB5 dataset is that existing evaluations included the recordings between gestures as a "no gesture" label, despite this causing large unresolved class imbalance as well as label leakage through padding operations and a lower distribution of recording duration. 

Recordings were resampled to the expected sampling frequency of each model and preprocessed and normalized according to the steps outlined in each model paper: (\emph{i}) \sys-EMG: 20-90 Hz band pass filtered; (\emph{ii}) TinyMyo: 20-90 Hz band pass filter (Butterworth, 4th order) followed by a 50 Hz notch filter and per-channel z-score normalization; (\emph{iii}) PhysioWave: No filtering, per-channel z-score normalization. The three models output an embedding per patch. For a fair comparison, we applied a uniform aggregation strategy across all three models: patch embeddings were concatenated across the channel dimension and averaged across time before being passed to the classification head. We added a single linear layer classification head and fine-tuned each pretrained model on the four downstream datasets. 

\paragraph{From scratch ablation study.} To quantify the impact of the pretraining phase on model performance, we also trained \sys-EMG from scratch with randomly initialized weights on each of the four downstream datasets, using the same subject level train/test split as the finetuning evaluation. The results in Table~\ref{tab:emg-ablation} demonstrate that pretraining consistently improved performance across all datasets. The NinaPro DB5 dataset exhibits the largest improvement, showing that pretraining the \sys-EMG foundation model is most valuable when labeled data is limited relative to task complexity, exactly the setting foundation models are designed to address.

\vspace{-0.1in}
\begin{table}[!h]
    \centering
    \caption{\sys-EMG ablations. \textbf{Bold} values indicate best performance per column. All evaluations are subject-independent. Metrics: Accuracy, F1, Balanced Accuracy (BAcc) and CLER (Classification Error Rate).}
    \resizebox{\columnwidth}{!}{%
      \begin{tabular}{l|cc|cc|cc|cc}
        \toprule
        & \multicolumn{2}{c|}{\textbf{Discrete Gestures}} & \multicolumn{2}{c|}{\textbf{EPN-612}} & \multicolumn{2}{c|}{\textbf{NinaPro DB5}} & \multicolumn{2}{c}{\textbf{UCI-EMG}} \\
        Tuning & BAcc\,$\uparrow$ & CLER\,$\downarrow$ & Acc & F1 & Acc & F1 & Acc & F1 \\
        \midrule
    From Scratch & 65.9 $\pm$ 0.6 & 34.3 $\pm$ 1.2 & 89.6 $\pm$ 0.2 & 89.7 $\pm$ 0.2 & 32.8 $\pm$ 1.8 & 30.6 $\pm$ 1.7 & 86.6 $\pm$ 1.9 & 86.5 $\pm$ 1.9 \\
    Full Finetune & \textbf{70.8 $\pm$ 0.6} & \textbf{27.6 $\pm$ 0.8} & \textbf{94.65 $\pm$ 0.4} & \textbf{94.7 $\pm$ 0.5} & \textbf{41.4 $\pm$ 2.8} & \textbf{38.8 $\pm$ 3.7} & \textbf{89.4 $\pm$ 1.1} & \textbf{89.3 $\pm$ 1.1} \\
        \bottomrule
      \end{tabular}%
    }
    \label{tab:emg-ablation}
\end{table}

\section{\sys vs. LaBraM: Downstream Task Performance}
\label{apx:downstream_codebook}

We compare the downstream task performance of \sys with LaBraM when both models are \textbf{pre-trained and fine-tuned on the same datasets} and with an identical training regime. We use a 10-fold subject-independent cross-validation to evaluate models on five classification benchmarks (Motor, Memory, Sleep, Eyes, ERP) as described in \S\ref{apx:downstream_finetuning_analysis}. Table~\ref{table: sota_codebook} shows that, when all training settings are equal, \sys achieves higher balanced accuracy than LaBraM on all tasks (up to 13\%), demonstrating the effectiveness of our tokenizer in training codebook-based models.

\vspace{-0.1in}
\begin{table}[!h]
  \caption{Comparison 
  of classification balanced accuracy between \sys and LaBraM,
  both pre-trained and finetuned on the same datasets 
  using identical settings. 
  }
  \centering
  \begin{tabulary}{\linewidth}{L|ccccc|c}
    \toprule
    Model & Motor & ERP & Memory & Sleep & Eyes &  Mean\\
    \midrule
    LaBraM (ours) &
    0.570  & 0.819  & 0.565 & 0.715 & 0.805 &  0.695\\
    \sys-EEG & 
    \textbf{0.700} & \textbf{0.876}  & \textbf{0.573} & \textbf{0.728} & \textbf{0.869} & \textbf{0.749}\\
    \bottomrule
  \end{tabulary}
  \label{table: sota_codebook}
\end{table}

\newpage
\section{Frozen Backbone Downstream Task Performance}
\label{apx:frozen_head_section}

For each foundation model baseline, we use the original architecture implementation and pretrained weights with the backbone frozen, adding only a trainable classification head
to the model's output embeddings / tokens according to their original implementations. CBraMod~\citep{wang2025cbramod} and NeuroGPT~\citep{NeuroGPT} employ a 3-layer MLP as their classification heads, while the remaining EEG and ECG models use a single layer. In EMG, we used 1-second input window with trainable classification heads. For \sys, we use all tokens produced by the multi-branch encoder as flattened input to a linear classification layer. The results
of classification-head-only training are reported in Tables ~\ref{tab:eeg_frozen}, ~\ref{tab:ecg_frozen} and ~\ref{tab:emg_frozen} using the evaluation schemes described in Section ~\ref{sec:downstream} and Appendix ~\ref{apx:downstream_finetuning_analysis}. For EMG, we evaluated on 1-second windows. 

\begin{table}[!h]
\centering
\caption{Frozen-backbone EEG classification results. Results are reported as mean balanced accuracy $\pm$ standard deviation. \textbf{Bold} and \underline{underlined} values indicate best and second-best performance per column, respectively. \\}
\label{tab:eeg_frozen}
\resizebox{0.8\textwidth}{!}{
\begin{tabular}{lccccc}
\toprule
Model & Motor & Memory & Sleep & Eyes & ERP \\
\midrule

BIOT
& 0.335 $\pm$ 0.039
& 0.528 $\pm$ 0.024
& ---
& 0.748 $\pm$ 0.046
& 0.500 $\pm$ 0.000 \\
CBraMod
& 0.440 $\pm$ 0.077
& 0.537 $\pm$ 0.016
& \underline{0.628 $\pm$ 0.044}
& 0.800 $\pm$ 0.034
& 0.500 $\pm$ 0.000\\
EEGPT
& 0.331 $\pm$ 0.041
& 0.541 $\pm$ 0.030
& \textbf{0.634 $\pm$ 0.043}
& \underline{0.815 $\pm$ 0.031}
& 0.523 $\pm$ 0.008 \\
NeuroGPT
& 0.452 $\pm$ 0.047
& 0.500 $\pm$ 0.000
& 0.304 $\pm$ 0.010
& 0.656 $\pm$ 0.048
& 0.501 $\pm$ 0.002\\
REVE
& \underline{0.574 $\pm$ 0.076}
& \underline{0.568 $\pm$ 0.030}
& 0.616 $\pm$ 0.029
& 0.761 $\pm$ 0.053
& \underline{0.728 $\pm$ 0.040}\\
LaBraM
& 0.327 $\pm$ 0.029
& 0.500 $\pm$ 0.000
& 0.303 $\pm$ 0.008
& 0.733 $\pm$ 0.051
& 0.500 $\pm$ 0.000 \\
\midrule
\sys-EEG
& \textbf{0.599 $\pm$ 0.044}
& \textbf{0.590 $\pm$ 0.033}
& 0.624 $\pm$ 0.028
& \textbf{0.823 $\pm$ 0.035}
& \textbf{0.802 $\pm$ 0.038} \\
\bottomrule
\end{tabular}
}
\end{table}

\vspace{-0.2in}
\begin{table}[!h]
\centering
\caption{Frozen-backbone ECG classification results. Results are reported as accuracy and balanced accuracy. \textbf{Bold} and \underline{underlined} values indicate best and second-best performance per column, respectively. \\}
\label{tab:ecg_frozen}
\resizebox{0.3\textwidth}{!}{
\begin{tabular}{lcc}
\toprule
Model
& \multicolumn{2}{c}{5-Class}\\
\cmidrule(lr){2-3}
& Acc. & Bacc \\
\midrule
HuBERT-ECG
& \underline{54.94}  & 35.55\\

ECGFounder
& \textbf{75.06} & \textbf{63.66} \\

\midrule
\sys-ECG
& 54.57 & \underline{42.13} \\
\bottomrule
\end{tabular}
}
\end{table}

\begin{table}[!h]
\centering
\caption{Frozen-backbone EMG classification results. Results are reported as mean accuracy/F1-score $\pm$ standard deviation. \textbf{Bold} and \underline{underlined} values indicate best and second-best performance per column, respectively. \\}
\label{tab:emg_frozen}
\resizebox{0.95\textwidth}{!}{
\begin{tabular}{lcccccc}
\toprule
Model
& \multicolumn{2}{c}{EPN-612}
& \multicolumn{2}{c}{NinaPro DB5}
& \multicolumn{2}{c}{UCI-EMG} \\
\cmidrule(lr){2-3}
\cmidrule(lr){4-5}
\cmidrule(lr){6-7}
& Acc. & F1 & Acc. & F1 & Acc. & F1 \\
\midrule
PhysioWave
& 89.36 $\pm$ 0.16
& 89.43 $\pm$ 0.16
& 18.07 $\pm$ 0.91
& 16.71 $\pm$ 1.42
& 84.21 $\pm$ 1.57
& 84.01 $\pm$ 1.58 \\

TinyMyo
& 75.28 $\pm$ 0.07
& 75.01 $\pm$ 0.09
& 21.60 $\pm$ 0.51
& 16.61 $\pm$ 0.49
& 83.67 $\pm$ 0.83
& 83.48 $\pm$ 0.93 \\
\midrule
\sys-EMG
& \textbf{90.38 $\pm$ 0.04}
& \textbf{90.41 $\pm$ 0.06}
& \textbf{25.54 $\pm$ 0.68}
& \textbf{21.91 $\pm$ 0.81}
& \textbf{88.45 $\pm$ 1.06}
& \textbf{88.45 $\pm$ 1.05} \\
\bottomrule
\end{tabular}
}
\end{table}

\newpage
\section{Tokenizer Swap Ablation}
\label{apx:swap_ablation}

We examine the effect of same backbone, parameter budget, pretraining data, normalization and downstream protocol but different tokenizers. We use two tokenizers from \S\ref{apx:unit-circle}, namely Row 2 in Table \ref{table:ablation} (4-Branch, 8CB, MSE Phase + MSE A) and Row 5 in Table \ref{table:ablation} (4-Branch, 8CB, Phase Loss + log(A) + Temporal), which share the same architecture and differ only in the training loss (val MSE 0.946 and 0.035 respectively). Using these two tokenizers, we pre-trained two \sys-EEG models on PhysioNet-MI~\citep{physionetmi} and benchmarked both on the High Gamma Motor dataset ~\citep{HighGamma}. We chose Motor because the tokenizers and consequently the foundation models are pre-trained on PhysioNet-MI (a motor dataset), and Motor is the downstream task most closely matched to that pretraining domain. Evaluation is under head-only (10-fold) and full-finetuning (5-fold) subject-wise cross-validation. In other words, \textbf{backbone architecture, parameter budget, pretraining data, normalization and downstream protocol are held identical. Only the tokenizer varies.} The results show that tokenization quality translates into downstream performance: balanced accuracy improves with tokenizer fidelity and the effect is clearest under full fine-tuning. 

\vspace{-0.1in}
\begin{table}[!h]
  \caption{Comparison of classification balanced accuracy between \sys-EEG on the Motor task pre-trained using various tokenizers.}
  \centering
  \begin{tabulary}{\linewidth}{L|cc}
    \toprule
    Tokenizer used to train same backbone model (corresponding val MSE) & Head-Only & Full-Fine Tuning\\
    \midrule
    4-Branch, 8CB, MSE Phase + MSE A (0.946) & 56.5 (8.59) & 58.5 (3.19) \\
    4-Branch, 8CB, Phase Loss + log(A) + Temporal (0.035) & \textbf{58.9 (5.67)} & \textbf{64.0 (1.39)} \\
    \bottomrule
  \end{tabulary}
  \label{table: swap_ablation_motor}
\end{table}

We extended this ablation study to other tokenizers from Section \S\ref{sec:scaling_analysis} in the EEG modality. To examine the relationship between number of codebooks and downstream performance, we used the tokenizers with (a) 4-Branch, 1CB (b) 4-Branch, 2CB and (c) 4-Branch, 8CB, which keeps the same architecture and training loss but scales only the number of codebooks.

\vspace{-0.1in}
\begin{table}[!h]
  \caption{Comparison of classification balanced accuracy between \sys-EEG on the Motor task pre-trained using various tokenizers.}
  \centering
  \begin{tabulary}{\linewidth}{L|cc}
    \toprule
    Tokenizer used to train same backbone model (corresponding val MSE) & Head-Only & Full-Fine Tuning\\
    \midrule
    4-Branch, 1CB (0.35) & 59.0 (4.91) & 62.7 (1.61)\\
    4-Branch, 2CB (0.12) & 59.1 (6.09) & 63.5 (2.23)\\
    4-Branch, 8CB (0.035) & 58.9 (5.67)	& 64.0 (1.39)\\
    \bottomrule
  \end{tabulary}
  \label{table: swap_ablation_branches}
\end{table}

We notice that once the tokenizer faithfully reconstructs task-relevant structure, further fidelity captures detail less relevant to the task-label with diminishing returns: a 10× MSE reduction (0.35→0.035) yields modest gain (saturation), whereas the low vs very good fidelity loss swap (0.946→0.035) gives a large 58.5→64.0. 

\newpage
\section*{NeurIPS Paper Checklist}

\begin{enumerate}

\item {\bf Claims}
    \item[] Question: Do the main claims made in the abstract and introduction accurately reflect the paper's contributions and scope?
    \item[] Answer: \answerYes{} 
    \item[] Justification: The abstract clearly outlines the core contributions presented in the main paper, particularly the introduction of the biosignal tokenizer family. The claims made are consistent with the scope and findings detailed throughout the work.

\item {\bf Limitations}
    \item[] Question: Does the paper discuss the limitations of the work performed by the authors?
    \item[] Answer: \answerYes{} 
    \item[] Justification: The discussion section describes how the architecture needs to be configured for each modality and how a unified (cross-modal) model can be challenging. In addition, it also outlines potential directions for future research, one of which being a better foundation model, a factor that might currently limit downstream task performance.

\item {\bf Theory assumptions and proofs}
    \item[] Question: For each theoretical result, does the paper provide the full set of assumptions and a complete (and correct) proof?
    \item[] Answer: \answerYes{} 
    \item[] Justification: Appendix A describes the theoretical assumptions and a complete and mathematically sound analysis of our proposed training loss.

    \item {\bf Experimental result reproducibility}
    \item[] Question: Does the paper fully disclose all the information needed to reproduce the main experimental results of the paper to the extent that it affects the main claims and/or conclusions of the paper (regardless of whether the code and data are provided or not)?
    \item[] Answer: \answerYes{} 
    \item[] Justification: The method presented in the paper is fully reproducible. All model parameters, datasets used and preprocessing / training steps are described in great detail. While one of the pre-training datasets used is self-collected, it constitutes only a minor portion of the data and does not significantly impact the reproducibility of the paper’s main claims or conclusions.

\item {\bf Open access to data and code}
    \item[] Question: Does the paper provide open access to the data and code, with sufficient instructions to faithfully reproduce the main experimental results, as described in supplemental material?
    \item[] Answer: \answerNo{} 
    \item[] Justification: We provide the main code and some of our models open-source. While all model details and all settings are thoroughly documented in the main paper and supplementary material, we do not provide full open access to all datasets. Specifically, one of the datasets is our own self-collected closed-access dataset. However, this dataset represents only a minor part of the pre-training data and does not materially affect the reproducibility of the main experimental results.

\item {\bf Experimental setting/details}
    \item[] Question: Does the paper specify all the training and test details (e.g., data splits, hyperparameters, how they were chosen, type of optimizer) necessary to understand the results?
    \item[] Answer: \answerYes{} 
    \item[] Justification: The paper provides comprehensive details on data preprocessing, hyperparameters and network settings. All relevant training and testing configurations are thoroughly described in the main paper and supplementary material.

\item {\bf Experiment statistical significance}
    \item[] Question: Does the paper report error bars suitably and correctly defined or other appropriate information about the statistical significance of the experiments?
    \item[] Answer: \answerYes{} 
    \item[] Justification: The paper reports statistical significance through cross-validation experiments, providing the mean and std across folds. These metrics are appropriate and sufficient to support the paper’s experimental claims.

\item {\bf Experiments compute resources}
    \item[] Question: For each experiment, does the paper provide sufficient information on the computer resources (type of compute workers, memory, time of execution) needed to reproduce the experiments?
    \item[] Answer: \answerYes{} 
    \item[] Justification: Yes, these details are provided in Section 5 and Appendices.
    
\item {\bf Code of ethics}
    \item[] Question: Does the research conducted in the paper conform, in every respect, with the NeurIPS Code of Ethics \url{https://neurips.cc/public/EthicsGuidelines}?
    \item[] Answer: \answerYes{}{} 
    \item[] Justification: All research was conducted responsibly and it fully adheres to the NeurIPS Code of Ethics. 

\item {\bf Broader impacts}
    \item[] Question: Does the paper discuss both potential positive societal impacts and negative societal impacts of the work performed?
    \item[] Answer: \answerYes{}{} 
    \item[] Justification: 
    The primary contribution of the paper is a biosignal tokenizer family. As discussed in the Introduction, Discussion and Conclusion, biosignal foundation models offer substantial application potential in neuroscience research, Brain-Computer Interfaces and healthcare. Therefore, NeuroRVQ advances the development of efficient large biosignal foundation models by providing a high-fidelity biosignal tokenizer, paving the way for broader use of foundation models in the biosignal space.
    
\item {\bf Safeguards}
    \item[] Question: Does the paper describe safeguards that have been put in place for responsible release of data or models that have a high risk for misuse (e.g., pre-trained language models, image generators, or scraped datasets)?
    \item[] Answer: \answerNA{} 
    \item[] Justification: We believe the method and data presented in the paper do not pose significant risks of misuse.

\item {\bf Licenses for existing assets}
    \item[] Question: Are the creators or original owners of assets (e.g., code, data, models), used in the paper, properly credited and are the license and terms of use explicitly mentioned and properly respected?
    \item[] Answer: \answerYes{} 
    \item[] Justification: All external assets used in this work are properly cited. 

\item {\bf New assets}
    \item[] Question: Are new assets introduced in the paper well documented and is the documentation provided alongside the assets?
    \item[] Answer: \answerYes{} 
    \item[] Justification: 
    We release the NeuroRVQ tokenizers, foundation models and code. The self-collected EEG dataset is not released due to intellectual property constraints. Its collection was approved by Imperial College London’s Research Governance and Integrity Team (approval number 22IC7505).

\item {\bf Crowdsourcing and research with human subjects}
    \item[] Question: For crowdsourcing experiments and research with human subjects, does the paper include the full text of instructions given to participants and screenshots, if applicable, as well as details about compensation (if any)? 
    \item[] Answer: \answerYes{} 
    \item[] Justification: Yes, the full text of the information provided to participants is included in Appendix F. Each participant received a £30 gift.

\item {\bf Institutional review board (IRB) approvals or equivalent for research with human subjects}
    \item[] Question: Does the paper describe potential risks incurred by study participants, whether such risks were disclosed to the subjects, and whether Institutional Review Board (IRB) approvals (or an equivalent approval/review based on the requirements of your country or institution) were obtained?
    \item[] Answer: \answerYes{} 
    \item[] Justification: The collection of the self-collected EEG dataset was approved by the Research Governance and Integrity Team of Imperial College London (approval number 22IC7505), the institutional equivalent of an IRB. The study involved non-invasive scalp EEG recordings during real and imagined movements and a movement-controlled game, which poses minimal risk. Participants were informed about the procedures and the use of their data before taking part and gave informed consent (Appendix F).
    

\item {\bf Declaration of LLM usage}
    \item[] Question: Does the paper describe the usage of LLMs if it is an important, original, or non-standard component of the core methods in this research? Note that if the LLM is used only for writing, editing, or formatting purposes and does \emph{not} impact the core methodology, scientific rigor, or originality of the research, declaration is not required.
    \item[] Answer: \answerNA{} 
    \item[] Justification: LLMs were used for writing and editing purposes

\end{enumerate}

\end{document}